\documentclass[letterpaper, 10 pt, conference]{ieeeconf}  % Comment this line out if you need a4paper

\IEEEoverridecommandlockouts                              % This command is only needed if 
\usepackage{graphics}
\usepackage{graphics} % for pdf, bitmapped graphics files
\usepackage{epsfig} % for postscript graphics files
\usepackage{mathptmx} % assumes new font selection scheme installed
\usepackage{times} % assumes new font selection scheme installed
\usepackage{amsmath} % assumes amsmath package installed
\usepackage{amssymb}  % assumes amsmath package installed
\usepackage{kotex}
\usepackage{times}
\usepackage{epsfig}
\usepackage{makecell} % 추가해야 함
\usepackage{graphicx}
\usepackage{inconsolata}
\usepackage{pifont}
\usepackage{dblfloatfix}
\usepackage{url}
\usepackage[table,xcdraw]{xcolor} 
\usepackage{graphicx} 
\usepackage{caption}
\usepackage{multirow}
\usepackage{float} 
\usepackage{placeins}
\usepackage{subfigure}
\usepackage{subcaption}
\usepackage{cite}

\definecolor{DarkYellow}{rgb}{0.87, 0.72, 0.0} % 원하는 RGB 값으로 정의
\definecolor{skyblue}{RGB}{0, 150, 255}  % 연한 하늘색
\definecolor{peach}{RGB}{255, 153, 102}    % 복숭아색 (PeachPuff)
\definecolor{lavender}{RGB}{148, 87, 235} % 진한 연보라색 (보라색 계열)
\definecolor{rose}{RGB}{255, 102, 102}

\newcommand{\xtiny}{\fontsize{3.5pt}{3.8pt}\selectfont} % \tiny보다 작은 폰트 크기 정의

\title{DroneKey: Drone 3D Pose Estimation in Image Sequences \\ using Gated Key-representation and Pose-adaptive Learning
}

\author{Seo-Bin Hwang$^{1}$ and Yeong-Jun Cho$^{1*}$% <-this % stops a space
\thanks{$^{1}$Department of AI Convergence, Chonnam National University, Korea.
        {\tt\small cnu.cvl.hsb@gmail.com and \tt\small yj.cho@jnu.ac.kr}}%
\thanks{*Corresponding Author}% <-this % stops a space
}

\makeatletter
\let\@oldmaketitle\@maketitle % 기존 \maketitle 저장
\renewcommand{\@maketitle}{\@oldmaketitle% 기존 타이틀 출력 후 추가
  \begin{center}
  \vspace{10pt}
    \parbox{\linewidth}{%
      \centering
      \includegraphics[width=0.95\linewidth]{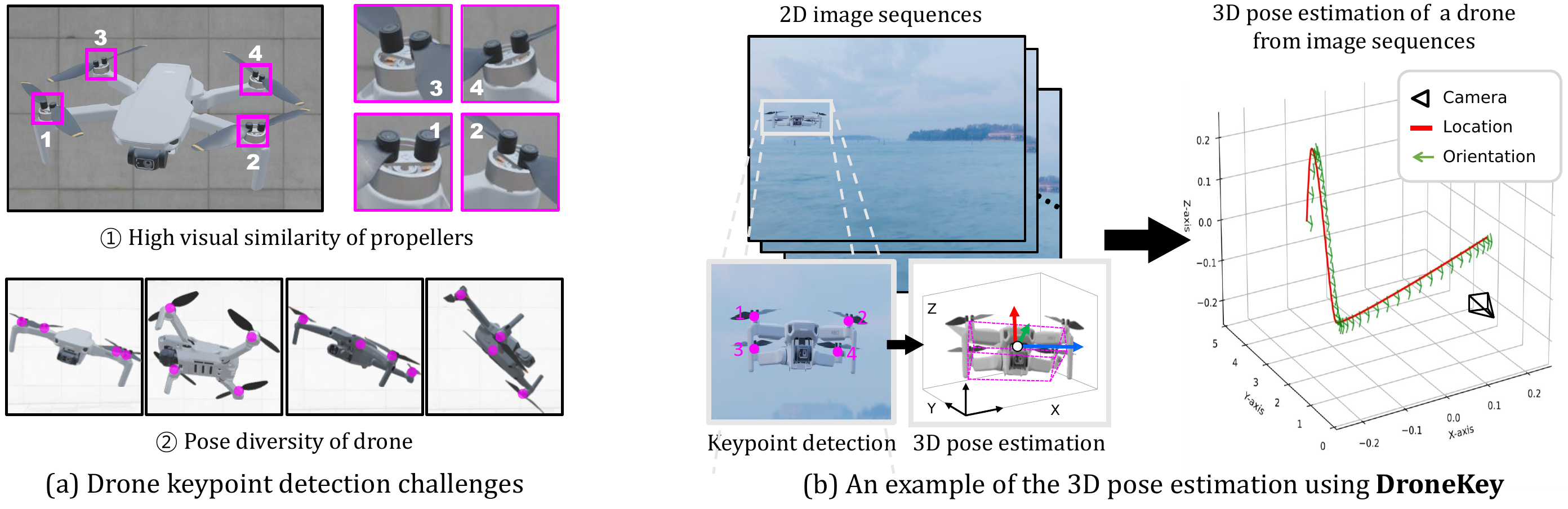}
      \captionof{figure}{\textbf{Overview of drone keypoint detection and 3D pose estimation.}
        (a) Challenges in drone keypoint detection.\\
        (b) Pipeline and result of 3D pose estimation from 2D image sequences through keypoint detection and 3D pose estimation.}
      \label{fig:01}
    }
  \end{center}
  \bigskip
  \vspace{-25pt}
}
\makeatother

\begin{document}

\maketitle
\thispagestyle{empty}
\pagestyle{empty}

\addtocounter{figure}{-1}
%■■■■■■■■■■■■■■■■■■■■■■■■■■■■■■■■■■■■■■■■■■■■■■■■■■■■■■■■■■■■■■■■■■■■■■■■■■■

\linespread{0.9} % 기본값은 1, 0.9로 줄이면 간격이 좁아짐

\begin{abstract}
Estimating the 3D pose of a drone is important for anti-drone systems, but existing methods struggle with the unique challenges of drone keypoint detection.
Drone propellers serve as keypoints but are difficult to detect due to their high visual similarity and diversity of poses.
% To overcome these challenges, we design a 2D keypoint detector, named DroneKey, specifically designed for drones.
To address these challenges, we propose DroneKey, a framework that combines a 2D keypoint detector and a 3D pose estimator specifically designed for drones.
% DroneKey extracts two key-representations (intermediate and compact) from each transformer encoder layer and optimally combines them using a gated sum. 
In the keypoint detection stage, we extract two key-representations (intermediate and compact) from each transformer encoder layer and optimally combine them using a gated sum. 
We also introduce a pose-adaptive Mahalanobis distance in the loss function to ensure stable keypoint predictions across extreme poses.
% 데이터를 만들었다 -> 뭘 하기 위해 -> 이걸 공개했다.
We built new datasets of drone 2D keypoints and 3D pose to train and evaluate our method, which have been publicly released.
Experiments show that our method achieves an AP of 99.68\% (OKS) in keypoint detection, outperforming existing methods.
Ablation studies confirm that the pose-adaptive Mahalanobis loss function improves keypoint prediction stability and accuracy.
Additionally, improvements in the encoder design enable real-time processing at 44 FPS.
For 3D pose estimation, our method achieved an MAE-angle of 10.62°, an RMSE of 0.221m, and an MAE-absolute of 0.076m, demonstrating high accuracy and reliability. The code and dataset are available at \textcolor{blue}{\url{https://github.com/kkanuseobin/DroneKey}}.
\end{abstract}

%■■■■■■■■■■■■■■■■■■■■■■■■■■■■■■■■■■■■■■■■■■■■■■■■■■■■■■■■■■■■■■■■■■■■■■■■■■■
%■■■■■■■■■■■■■■■■■■■■■■■■■■■■■■■■■■■■■■■■■■■■■■■■■■■■■■■■■■■■■■■■■■■■■■■■■■■
\linespread{0.85} % 기본값은 1, 0.9로 줄이면 간격이 좁아짐
\section{INTRODUCTION}
\label{sec:introduction}

In recent years, drones have been widely used across various domains, including videography, delivery, and military applications. However, their misuse for illegal purposes has also been increasing.
To address this issue, an anti-drone system\textcolor{blue}{\cite{anti-drone}} has been proposed, which estimates a drone's 3D pose (i.e., position and orientation).
Estimating a drone's 3D pose also allows for determining its gaze direction and position, which are crucial for assessing its potential illegality.

% ♣ 여기 빠졌음!
Various studies have explored estimating a drone's 3D pose from single-camera images.
For example, G. Albanis et al.\textcolor{blue}{\cite{dronepose-sil}} proposed a 3D pose estimation method based on silhouette analysis.
However, due to the symmetrical drone appearance, this method cannot accurately estimate drone orientations.
To address this issue, previous studies\textcolor{blue}{\cite{drone-mask, drone-simcc}} have explored detecting 2D keypoints as a more reliable approach for 3D pose estimation.
This approach is effective because keypoints represent the drone's 3D pose and allow for 3D pose reconstruction using geometric computations (e.g., PnP solver\textcolor{blue}{\cite{pnp}}) as depicted in Fig. \textcolor{blue}{\ref{fig:01}} (b).

% ♣ 여기 빠졌음!
However, detecting 2D keypoints from drone images presents two challenges.
First, drone keypoints, particularly propellers, exhibit high visual similarity, as shown in Fig. \textcolor{blue}{\ref{fig:01}} (a)-①.
The high visual similarity of propellers makes it difficult to determine their order and arrangement.
In keypoint detection, it is crucial not only to locate keypoints but also to infer their correct order.
Second, drones have diverse poses, as shown in Fig. \textcolor{blue}{\ref{fig:01}} (a)-②.
Unlike humans or vehicles, drones can freely rotate in the air.
In the case of humans, keypoint positions are predictable; for example, the head is above the shoulders, and both legs are positioned lower.
However, for drones, keypoints are arranged more freely, making it difficult to predict their position and order.

In this study, we employ a transformer\textcolor{blue}{\cite{transformer}} structure as the baseline for the proposed DroneKey to leverage its ability to capture both spatial relationships and keypoint orders.
Since transformers tend to focus on object-level features\textcolor{blue}{\cite{obj-level}}, we need to extract compact representations of keypoint shapes to enhance spatial locality.
Moreover, to efficiently train the large number of network parameters in the transformer with a limited amount of training data, we maximized the use of mid-level features from the transformer encoder.
To this end, we propose a novel keypoint head with the following approach:
by utilizing all encoder layers, we incorporate diverse features to enhance keypoint detection.
We employ gated sum integration to learn the importance of each layer, allowing for optimal weight assignment and more effective training.
As a result, keypoints for all propellers with similar visuals can be accurately detected in the correct order.

Furthermore, to improve keypoint detection in extreme drone poses, we propose a pose-adaptive Mahalanobis distance for our loss function.
To this end, we analyze the arrangement of drone keypoints by computing the covariance matrix, which represents their alignment according to pose variations.
Based on the covariance matrix, we compute the Mahalanobis distance in our loss function to adaptively assess each keypoint error relative to the drone’s pose.
To enhance the stability of the loss function, we design the loss function to be adjustable using the proposed scale function over the training time.

To validate the proposed methods, we provide new drone datasets, such as \texttt{2DroneKey} and \texttt{3DronePose}, and extensively evaluate them by comparing them with other state-of-the-art methods.
Our proposed DroneKey offers a pipeline from 2D keypoint detection to 3D pose estimation in image sequences, as shown in Fig. \textcolor{blue}{\ref{fig:01}} (b).
The main contributions of this study are as follows:
% \vspace{-5pt}
\begin{itemize}
    \item \textbf{First study on drone-specific keypoint detection}: Unlike prior works using general methods, we propose a framework tailored to drone keypoint detection.
    \item \textbf{Provide a new dataset}: We created and released a synthetic dataset using real-world 360-degree footage as backgrounds to address the lack of drone keypoint and 3D pose datasets.
    \item \textbf{Real-time performance}: With an optimized architecture and 6 encoder layers, our model achieves 44 FPS.
    \item \textbf{High accuracy}: Our method achieved an AP of 99.68\% (OKS)\textcolor{blue}{\cite{oks}}. The estimated 3D pose showed minimal errors, achieving an MAE-angle of 10.62°, an RMSE of 0.221m, and an MAE-absolute of 0.076m.
\end{itemize}
% \vspace{-5pt}
These results enable 3D drone pose estimation from 2D images, making the proposed method suitable for anti-drone applications requiring 3D pose information.

%■■■■■■■■■■■■■■■■■■■■■■■■■■■■■■■■■■■■■■■■■■■■■■■■■■■■■■■■■■■■■■■■■■■■■■■■■■■
%■■■■■■■■■■■■■■■■■■■■■■■■■■■■■■■■■■■■■■■■■■■■■■■■■■■■■■■■■■■■■■■■■■■■■■■■■■■
\vspace{10pt}
\section{RELATED WORKS}
\label{sec:relatedworks}

% ♣ 여기 빠졌음!
The objectives of this study are 2D keypoint detection in image sequences and 3D pose estimation of the drone.
Drone 3D pose estimation methods using 2D images and keypoint detection methods are reviewed in this section.

\begin{figure*}[t]
    % \vspace{-10pt}
	\centering
	\includegraphics[width=0.97\linewidth]{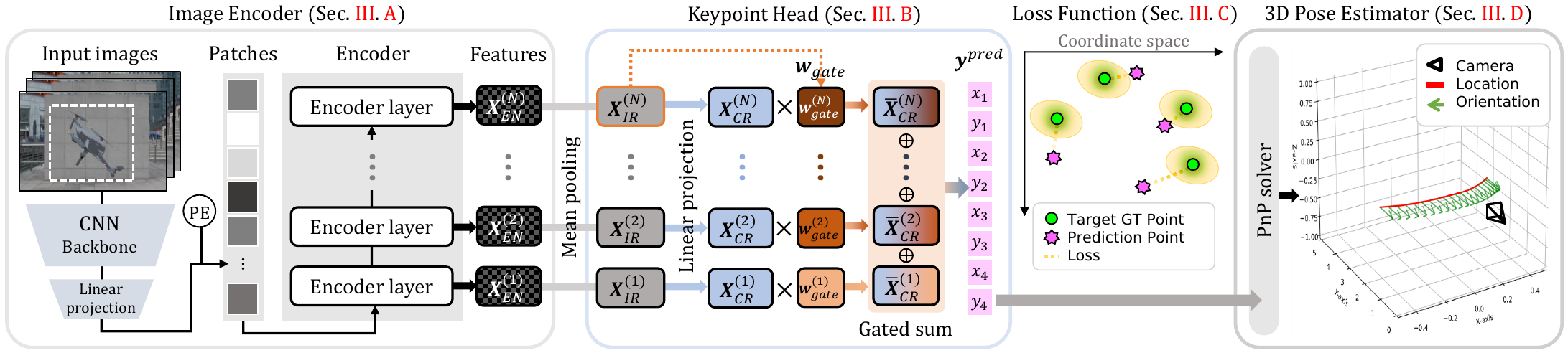}
	\caption{\textbf{Overall framework for drone keypoint detection and 3D pose estimation.}}
	\label{fig:02}
    \vspace{-15pt}
\end{figure*}

\subsection{Drone 3D Pose Estimation using 2D Images} 

Estimating a drone’s 3D pose (i.e., 3D position and orientation) using only a single camera is challenging.
From a geometric perspective, at least two cameras or depth-based sensors are required to determine 3D pose.
Nevertheless, estimating 3D pose using only a single RGB camera is highly practical, making it a widely studied topic in recent research.
Fu et al.\textcolor{blue}{\cite{drone-marker}} proposed a drone pose estimation based on tracking markers attached to the drone.
However, this method requires known markers on the drone, which limits its practicality.
Albanis et al.\textcolor{blue}{\cite{dronepose-sil}} estimated the 3D pose of a drone using its silhouette mask without requiring markers.
However, this method easily confuses the drone's orientation due to the drone's symmetric appearance.
These methods lack practicality and accuracy.

To address these issues, many studies have focused on extracting keypoints from 2D images and estimating 3D pose based on the extracted keypoints.
For drones, the primary keypoints are the propellers. Since their arrangement directly reflects the drone’s orientation and position.
To this end, Jin et al.\textcolor{blue}{\cite{drone-mask}} proposed drone keypoint detection by using mask R-CNN\textcolor{blue}{\cite{maskrcnn}}.
You et al.\textcolor{blue}{\cite{drone-simcc}} proposed a simple coordinate classification (SimCC)\textcolor{blue}{\cite{simcc}} for drone keypoint detection.
These studies\textcolor{blue}{\cite{drone-mask, drone-simcc}} merely applied existing object detection methods for drone keypoint detection.
However, they did not fully account for the challenges of drone keypoint extraction as shown in Fig. \textcolor{blue}{\ref{fig:01}} (a).

\subsection{Keypoint Detection}
% ♣ 여기 빠졌음!
Keypoint detection has been studied for 2D pose estimation of objects.
Since keypoints are represented in coordinate space, they are typically predicted using coordinate regression.
Toshev et al.\textcolor{blue}{\cite{deeppoose}} proposed a keypoint detection using deep neural network regression of keypoint coordinates.
However, this keypoint regression approach is challenging because it requires predicting the exact position of a keypoint within an $H \times W$ image.
This means the model must determine a precise coordinate for all possible keypoint positions.
Its limitation is failing to model probabilistic distributions, making it sensitive to pixel shifts and ambiguous keypoints.

To address this issue, Tompson et al.\textcolor{blue}{\cite{heatmap-joint}} proposed a heatmap regression model.
By regressing on a ground truth heatmap, this approach handles keypoint uncertainty and improves learning in high-dimensional space.
In another study, Xu et al.\textcolor{blue}{\cite{vitpose}} enhanced keypoint detection by using vision transformer (ViT)\textcolor{blue}{\cite{vit}} as a backbone.
Similarly, Li et al.\textcolor{blue}{\cite{tokenpose}} propose a method that generates tokens containing keypoint heatmaps.
However, heatmap regression struggles to accurately generate the ground-truth heatmap at low-resolution images.
While higher resolutions improve accuracy, they also increase computational cost.

We reviewed studies on drone 3D pose estimation and keypoint detection.
Existing methods did not fully consider keypoint detection challenges (Fig. \textcolor{blue}{\ref{fig:01}} (a)).
Keypoint detection methods lack consistency due to pixel sensitivity and resolution dependence.
To address these issues, we used a transformer network to learn relationships between keypoints.
In addition, we designed a novel pose-adaptive Mahalanobis loss to address challenges in drone keypoint detection and enhance training efficiency.
This proposed loss integrates the benefits of both coordinate regression and heatmap regression.

%■■■■■■■■■■■■■■■■■■■■■■■■■■■■■■■■■■■■■■■■■■■■■■■■■■■■■■■■■■■■■■■■■■■■■■■■■■■
%■■■■■■■■■■■■■■■■■■■■■■■■■■■■■■■■■■■■■■■■■■■■■■■■■■■■■■■■■■■■■■■■■■■■■■■■■■■

\section{PROPOSED METHODS} 
\label{sec:methods} 
% ♣ 여기 빠졌음!
The proposed 3D drone pose estimation framework, named DroneKey, consists of four components, as shown in Fig. \textcolor{blue}{\ref{fig:02}}.
It detects the four propellers of the drone as keypoints for estimating the drone's 3D pose.
Then, the drone's 3D pose is geometrically computed using a PnP solver.

    %□□□□□□□□□□□□□□□□□□□□□□□□□□□□□□□□□□□□
    \subsection{Image Encoder for Extracting Representative Features}
    CNN-based methods\textcolor{blue}{\cite{maskrcnn,cnn-01,cnn-02}} mainly focus on learning local visual features, limiting their ability to capture relationships between keypoints.
    However, as described in Sec. I, the drone's keypoints (i.e., propellers) have the same appearance. 
    While CNN-based methods can detect individual propellers, they cannot predict their spatial arrangement for pose estimation.
    %(yjcho) 윗 부분은 introduction작성이후 다시 정리 필요 (이부분은 proposed가 아니라 motivation이나 main idea로..)
    To handle this issue, we employ a transformer encoder\textcolor{blue}{\cite{transformer}} with self-attention to learn global keypoint relationships.
    A CNN backbone extracts a feature $\mathbf{X}$ from the input image $\mathbf{I} \in \mathbb{R}^{H \times W \times C}$, where $H$ and $W$ denote the height and width of the image, respectively, and $C$ represents the number of channels.
    The extracted feature from the CNN backbone can be represented as $\mathbf{X} \in \mathbb{R}^{\frac{H}{32} \times \frac{W}{32} \times d}$, where $d$ denotes the feature dimension.
    The feature is simply tokenized as $\mathbf{x}_p \in \mathbb{R}^d$, where $p = 1,2,...,P$, denotes the index of the visual token, and $P=\frac{H}{32} \times \frac{W}{32}$.
    
    Before inputting the visual token $\mathbf{x}_p$ into the transformer encoder, positional encoding ($\mathrm{PE}$) is added to preserve spatial information by
    $\mathbf{x}_p^{(0)} = \mathrm{PE}(\mathbf{x}_p) + \mathbf{x}_p$.
    A set of input visual tokens $\mathbf{X}^{(0)}=\{\mathbf{x}_1^{(0)},\mathbf{x}_2^{(0)},...,\mathbf{x}_p^{(0)}\}$ is then processed through the self-attention transformer layers as follows:
    \begin{equation}
    \small
    \mathbf{X}^{(l)} = \mathrm{selfAttention}(Q=\mathbf{X}^{(l-1)},K=\mathbf{X}^{(l-1)},V=\mathbf{X}^{(l-1)}),
    \end{equation}
    where $Q$, $K$, $V$ are query, key, and value for self-attention inputs, respectively, and $l=1,2,...,N$ represents the index of the self-attention layer.
    We follow the transformer encoder structure as proposed in\textcolor{blue}{\cite{transformer}}. It stacks $N$ self-attention layers with fully connected layers~(FCN) to construct the transformer encoder, enabling it to capture visual relationships between input visual tokens.
    Thus, the output of each encoder can be represented by $\mathbf{X}_{EN}^{(l)}=FCN(\mathbf{X}^{(l)})$.

    %□□□□□□□□□□□□□□□□□□□□□□□□□□□□□□□□□□□
    \subsection{keypoint Head for Detecting 2D Keypoint Positions}
    
    The final output of the image encoder, $\mathbf{X}_{EN}^{(N)}$, contains high-level contextual information learned through multiple self-attention layers.
    In general, only the output from the final layer $\mathbf{X}_{EN}^{(N)}$ is used. 
    However, inspired by previous works\textcolor{blue}{\cite{SegFormer, EGTR}} that leveraged mid-level feature representations, we utilize all feature representations, i.e., $\{\mathbf{X}_{EN}^{(1)},\mathbf{X}_{EN}^{(2)},...,\mathbf{X}_{EN}^{(N)}\}$.
    
    To utilize all the mid-level feature representations in the keypoint heads, we further transform them as follows.
    First, we apply mean pooling to each feature representation along the feature dimension.
     \begin{equation} 
        \mathbf{X}_{IR}^{(l)} = \frac{1}{P}\sum_{p=1}^{P} \mathbf{x}^{(l)}_{p},
    \end{equation}
    % ♣ 여기 빠졌음! -> important
    where $\mathbf{x}^{(l)}_{p}$ is the $p$-th visual token of $l$-th encoder layer output $\mathbf{X}_{EN}^{(l)}$.
    We call this averaged feature representation intermediate representation$(IR)$. 
    Then, $\mathbf{X}_{IR}^{(l)}\in\mathbb{R}^d$ aggregates information from multiple patches into a unified representation, summarizing the features learned at each layer.

    Second, we further apply linear projection to $\mathbf{IR}^{(l)}$ as follows:
    \begin{equation} 
        \mathbf{X}_{CR}^{(l)} = \mathbf{W} \cdot \mathbf{X}_{IR}^{(l)}+\mathbf{b}, 
    \end{equation}
    % ♣ 여기 빠졌음! -> total
    where $\mathbf{W} \in \mathbb{R}^{2K \times d}$ is a trainable weight matrix and $\mathbf{b}$ is a bias vector.
    Here, $K$ represents the number of keypoints in drones and $2K$ corresponds to the two-dimensional coordinates $(x,y)$ of these keypoints.
    We set $K=4$ since the keypoints correspond to the drone propellers.
    We call the $2K$-dimensional features the compact representation (CR), which is aligned with the dimension of the ground truth coordinate values.
       
    To effectively aggregate the $\mathbf{X}_{CR}^{(l)}$ extracted from each encoder layer, we employ a gated summation method. 
    To this end, we generate a weight vector from the intermediate representation of the final layer, $\mathbf{X}_{IR}^{(N)}$, by applying a linear projection and softmax function, as follows:
    \begin{equation} 
        \mathbf{w}_{gate} = \mathrm{softmax}(\mathbf{W}^g \cdot \mathbf{X}_{IR}^{(N)} + \mathbf{b}^g), 
    \end{equation} 
    where $\mathbf{W}^g \in \mathbb{R}^{N \times d}$ is a trainable weight matrix and $\mathbf{b}^g$ is a bias vector. 
    Then, the $N$-dimensional weight vector $\mathbf{w}_{gate}$ is element-wise multiplied with each layer’s $\mathbf{X}_{CR}^{(l)}$ as follows:
    \begin{equation} 
        \bar{\mathbf{X}}_{CR}^{(l)} = \mathbf{w}^{(l)}_{gate} \mathbf{X}_{CR}^{(l)}, 
    \end{equation} 
    where $\mathbf{w}^{(l)}_{gate}$ denotes $l$-th weight value in the $\mathbf{w}_{gate}$.
    We then aggregate the weighted compact representations by $\mathbf{X}_G = \sum^{N}_{l = 1} \bar{\mathbf{X}}_{CR}^{(l)}$.
    This aggregated representation effectively integrates crucial information extracted from all layers, resulting in a richer and more informative representation.
    Finally, the gated summation result, $\mathbf{X}_G \in \mathbb{R}^{2K}$, is passed through the rectified linear unit (ReLU) function to predict the final keypoint coordinates as the following equation.
    \begin{equation} 
        \mathbf{y}^{pred} = \mathrm{ReLU}(\mathbf{X}_G), 
    \end{equation}
    where the ReLU function sets negative values to zero, enabling the model to learn stable key-representations and preventing unrealistic negative coordinate predictions.

        %□□□□□□□□□□□□□□□□□□□□□□□□□□□□□□□□□□□□

        \begin{figure}[t]
    	\centering
        \vspace{5pt}
            %\captionsetup{font=footnotesize, skip=2pt} % 글씨 크기 조절 및 자간 축소
            \subfigure[case 1] 
            {\includegraphics[width=0.32\columnwidth]{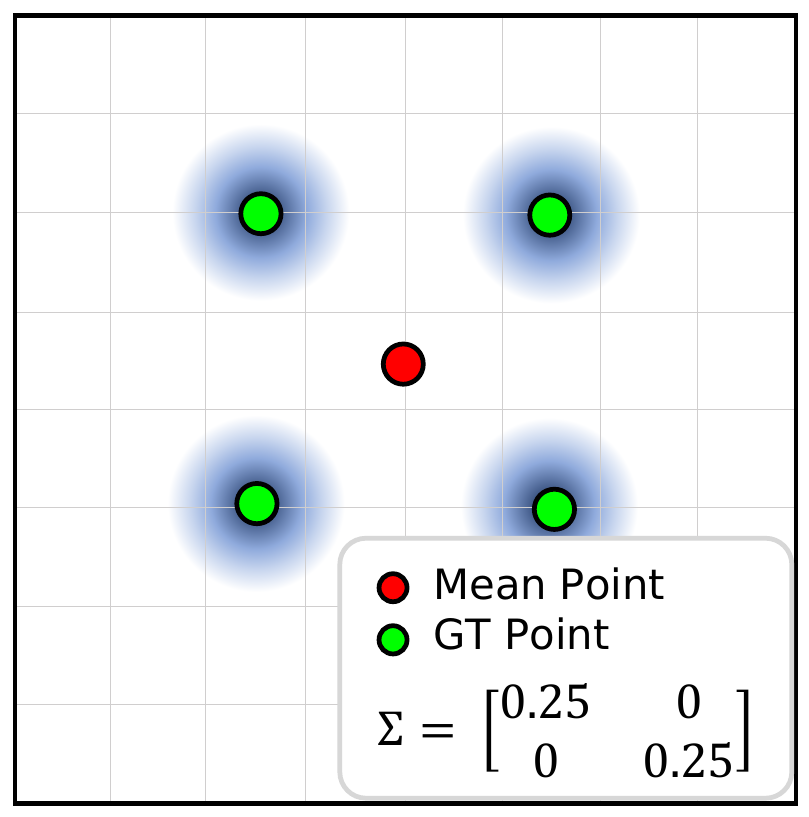}}
            \subfigure[case 2] 
            {\includegraphics[width=0.32\columnwidth]{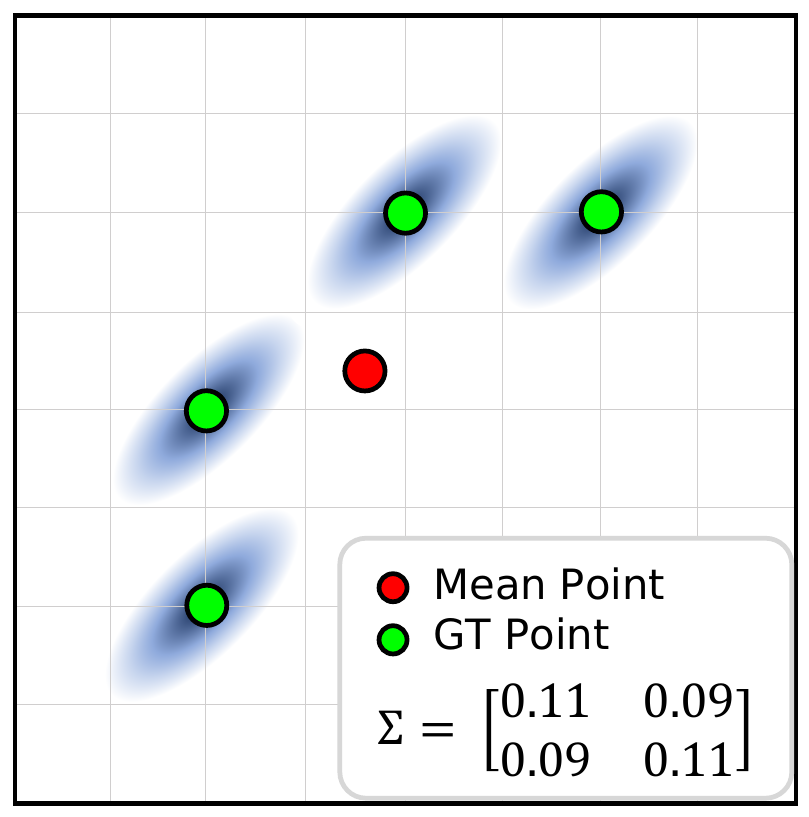}}
            \subfigure[case 3] 
            {\includegraphics[width=0.32\columnwidth]{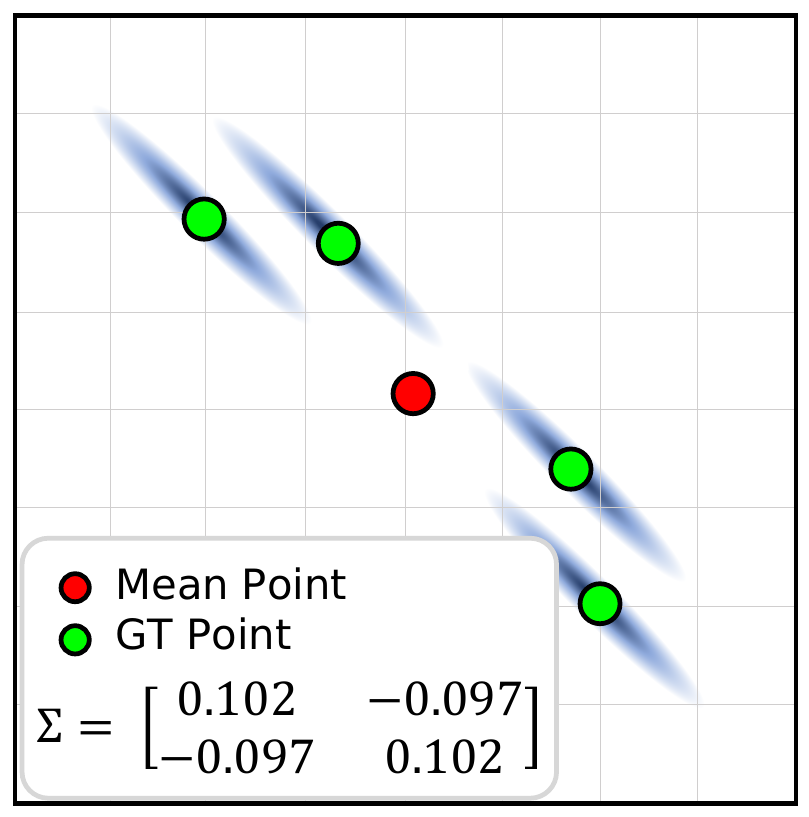}} \vspace{-5pt}
            \caption{\textbf{Distributions of keypoints and their covariance matrices.}
            % (a-b) The scatter matrix is computed from GT keypoints to derive principal axes (eigenvectors).
            % (c) Gaussian distributions are adapted to the pose using the covariance structure.
            % Dashed lines (\textcolor{DarkYellow}{- -}) represent Gaussian distributions adjusted per GT keypoint based on drone pose.
            }
                \vspace{-5pt}
    	\label{fig:03}
            \end{figure}

        \begin{figure}[t]
    	\centering
            \subfigure[$S(t) = 10$] {\includegraphics[width=0.32\columnwidth]{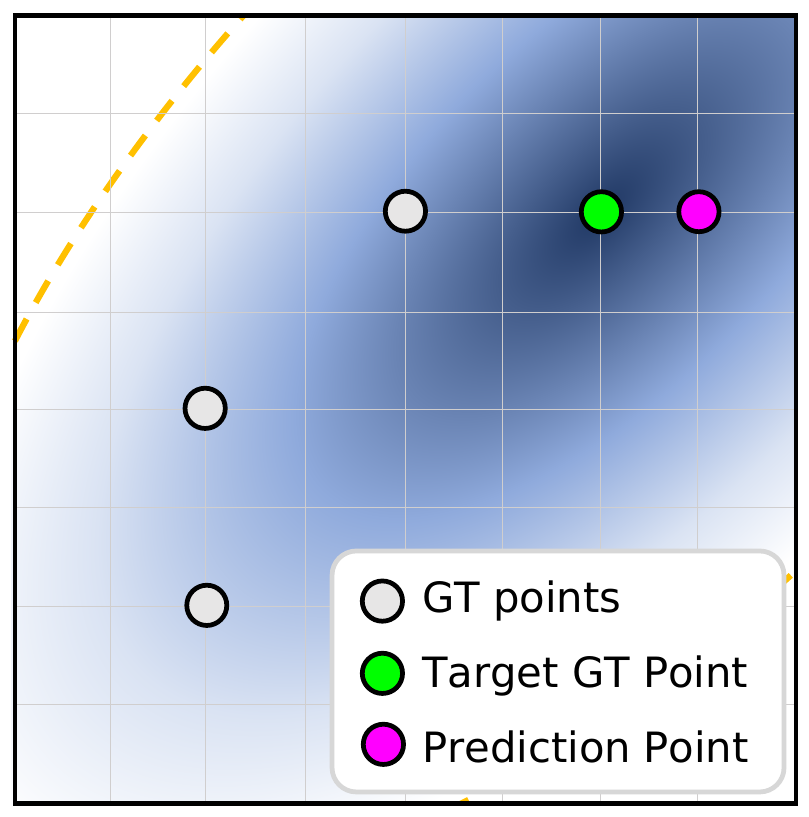}}
            \subfigure[$S(t) = 7$] {\includegraphics[width=0.32\columnwidth]{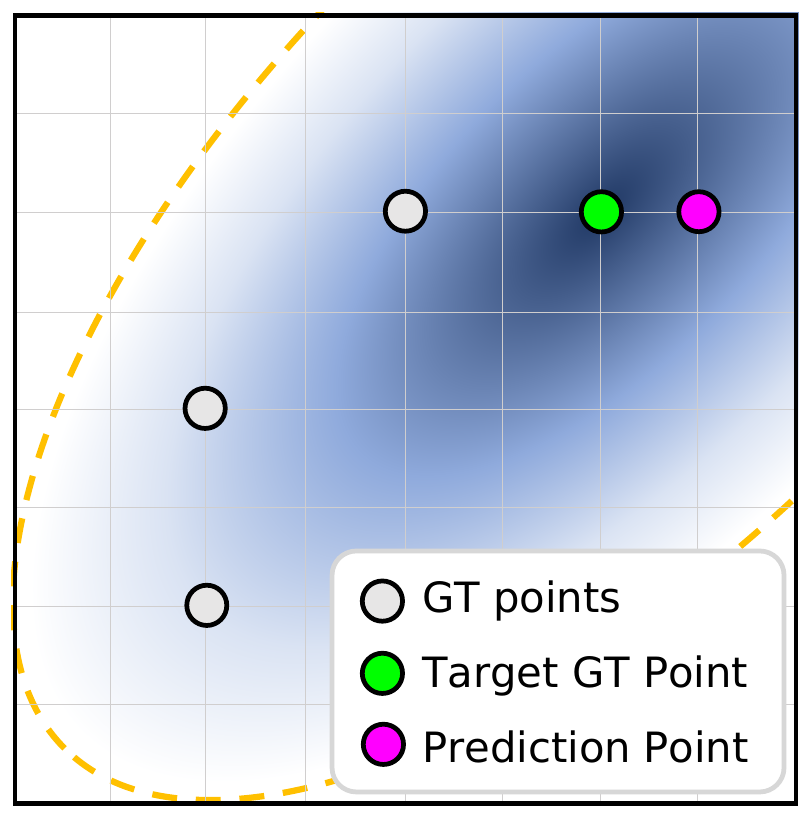}}
            \subfigure[$S(t) = 6$] {\includegraphics[width=0.32\columnwidth]{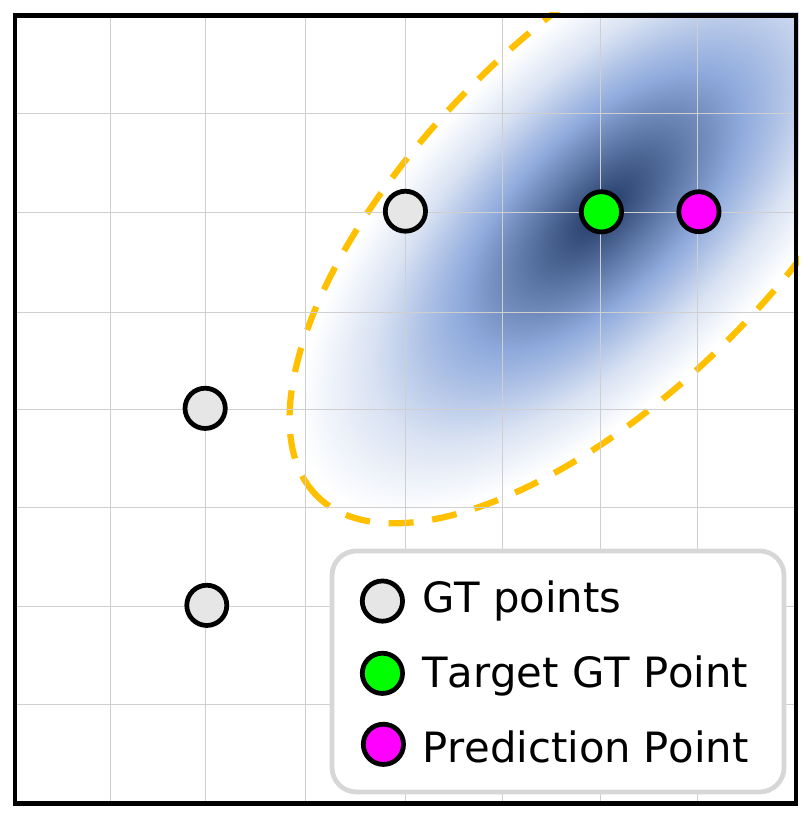}}
            \subfigure[$S(t) = 1.8$] {\includegraphics[width=0.32\columnwidth]{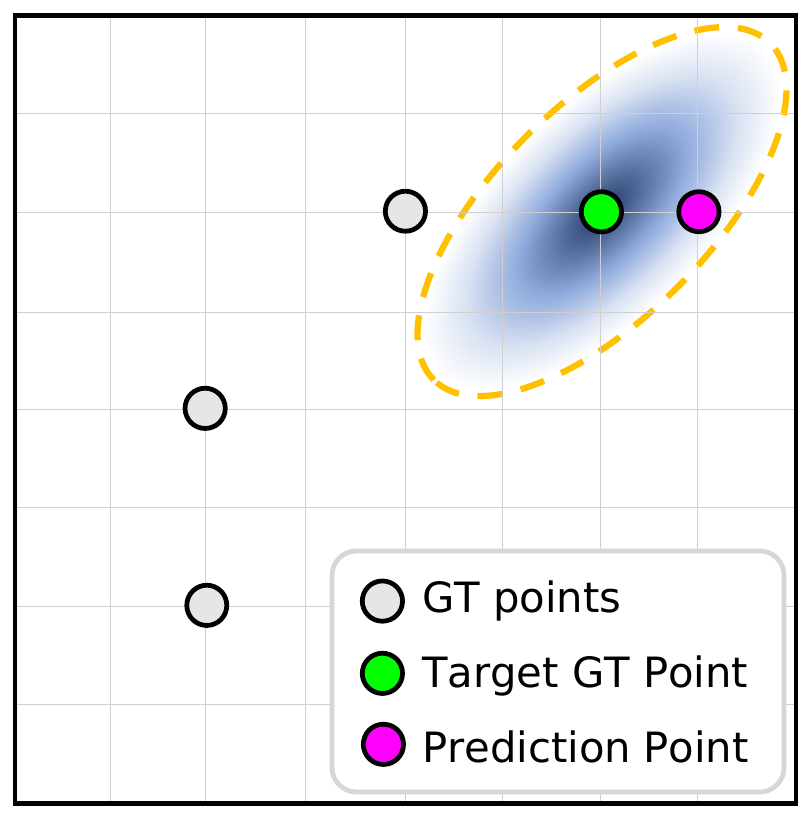}}
            \subfigure[$S(t) = 1$] {\includegraphics[width=0.32\columnwidth]{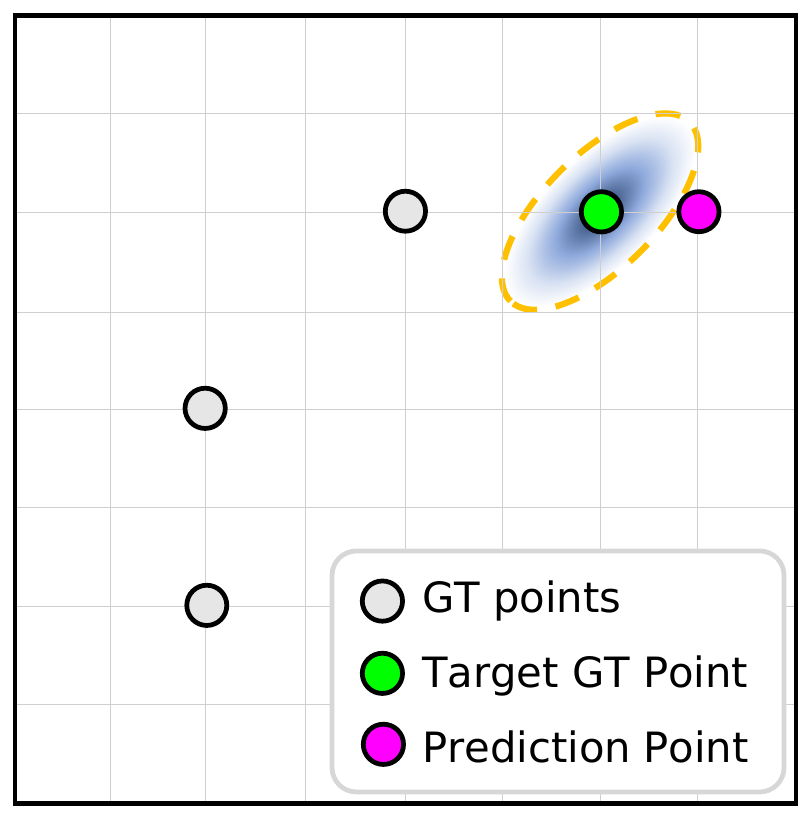}}
            \subfigure[$S(t) = 0.5$] {\includegraphics[width=0.32\columnwidth]{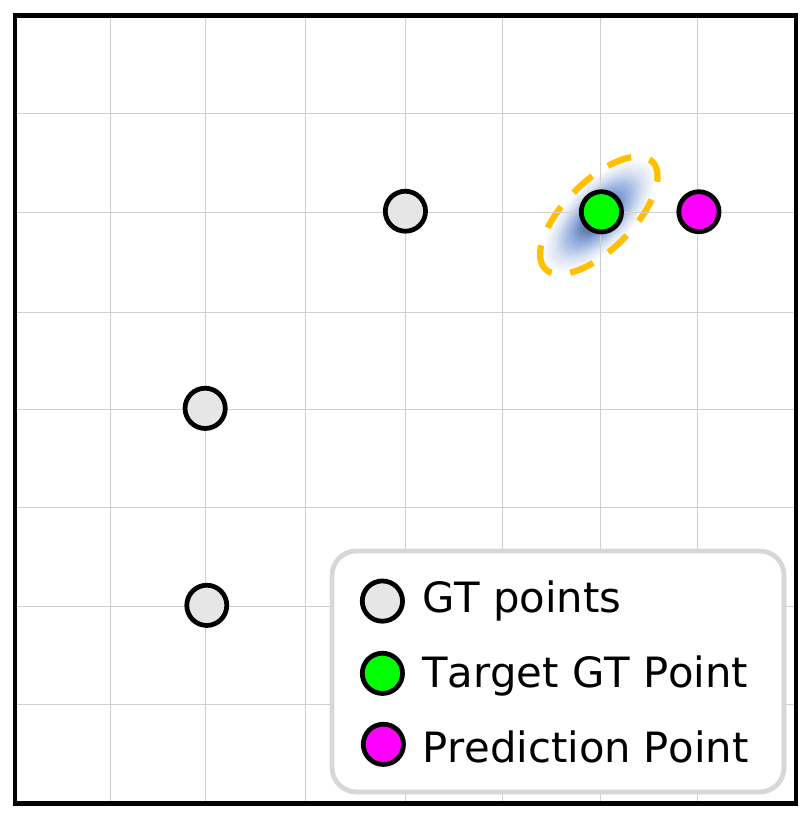}} \vspace{-5pt}
    	\caption{\textbf{Effect of scale values in pose-adaptive Mahalanobis distance.}  
                As the scale value increases, the allowable range expands, whereas a narrower distribution leads to convergence toward a stricter loss. % (\textcolor{DarkYellow}{- -}) represents the variance of the Gaussian distribution, which adapts to the pose.
                } \vspace{-10pt}
    	\label{fig:04}
        \end{figure}

        \begin{figure}[t]
        \vspace{5pt}
        	\centering
        	\includegraphics[width=0.985\linewidth]{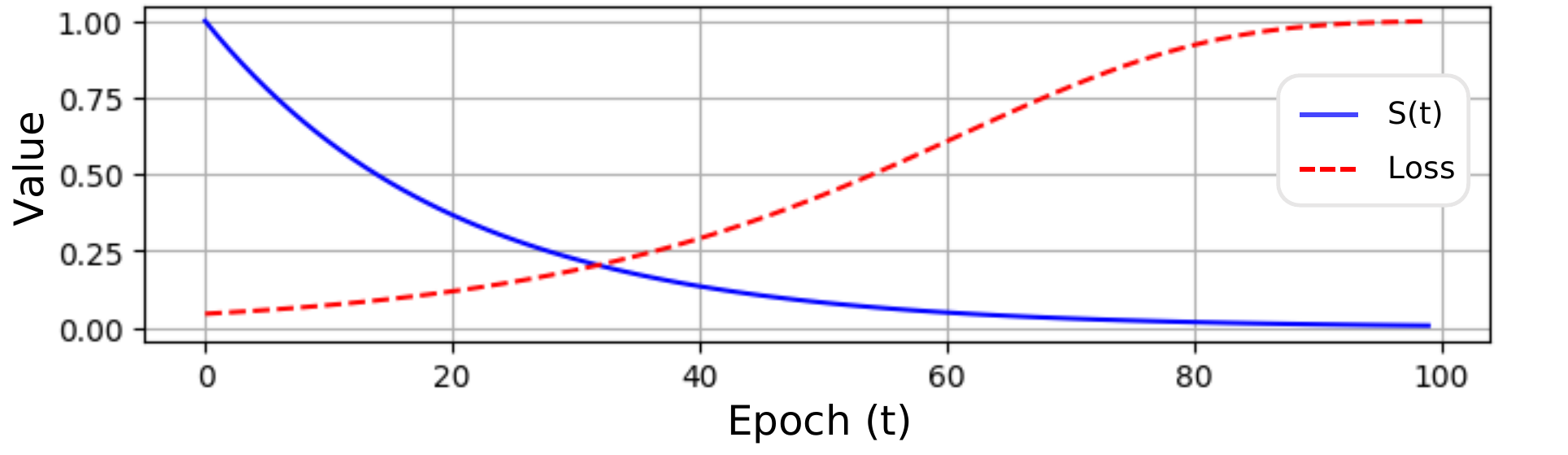}
            \vspace{-5pt}
        	  \caption{\textbf{Scale value and loss over epochs~($t$).}  
                        The plot illustrates the evolution of scale value (\textcolor{blue}{--}) and loss (\textcolor{red}{- -}) over training epochs.}
                        % Although the difference between GT and prediction remains at approximately (0.1, 0),  
                        % the decreasing sigma demonstrates a progressively more precise evaluation as training progresses.}
                \label{fig:05}
                \vspace{-15pt}
        \end{figure}

\begin{table*}[b]
\vspace{-5pt}
\centering
\caption{\textbf{Datasets}. The dataset is divided into \texttt{2DroneKey} for keypoint detection and \texttt{3DronePose} for pose estimation.}
\label{tab:01}
\tiny
\resizebox{\textwidth}{!}{ % 전체 너비 확장
\xtiny
    \renewcommand{\arraystretch}{1} % 이 테이블에서만 적용
    \begin{tabular}{c||cc|ccc}
    \hline \noalign{\hrule height 0.2pt}
    \multirow{2}{*}{Sequences} & \multicolumn{2}{c|}{ \texttt{2DroneKey}}   & \multicolumn{3}{c}{\texttt{3DronePose}}                \\ \cline{2-6} 
                    & \multicolumn{1}{c|}{Sequences~1 to 5} & Sequences~6 to 10 & \multicolumn{1}{c|}{Sequence~11} & \multicolumn{1}{c|}{Sequence~12} & \multicolumn{1}{c}{Sequence~13} \\ \hline
        Usage       & \multicolumn{2}{c|}{keypoint detection train, valid and test}  & \multicolumn{3}{c}{3D pose estimation validation}  \\ \hline
        Drone type  & \multicolumn{1}{c|}{Air2S}  & Mini2   & \multicolumn{3}{c}{Mini2} \\ \hline
        Frame       & \multicolumn{2}{c|}{1,000 frames per sequence} & \multicolumn{1}{c|}{500 frames} & \multicolumn{1}{c|}{400 frames} & \multicolumn{1}{c}{300 frames} \\ \hline
        Translation & \multicolumn{2}{c|}{}                                                & \multicolumn{1}{c|}{\ding{52}} & \multicolumn{1}{c|}{\ding{52}} & \multicolumn{1}{c}{\ding{52}} \\ \hline
        Rotation    & \multicolumn{2}{c|}{\ding{52}}  & \multicolumn{1}{c|}{}          & \multicolumn{1}{c|}{}       & \multicolumn{1}{c}{\ding{52}}\\ \hline 
        Nonlinear translation& \multicolumn{2}{c|}{}                 & \multicolumn{1}{c|}{} & \multicolumn{1}{c|}{\ding{52} }& \multicolumn{1}{c}{\ding{52}} \\ \hline\noalign{\hrule height 0.2pt}
    \end{tabular}
} % \resizebox 끝
\end{table*}

    \subsection{Pose-adaptive Mahalanobis Loss} 
    To optimize the networks, including the encoder (Sec. III.A) and keypoint heads (Sec. III.B), for accurate keypoint detection in drones, we considered two  issues when designing the loss function.
    First, the drone exhibited pose diversity, requiring keypoint prediction to be adaptive to variations in pose.
    Second, the proposed loss must converge stably and achieve a good solution.
    Thus, our loss function accounted for both aspects: adaptability and stability for the drone keypoint detection problem. 

    For adaptability, we design a loss function that dynamically adjusts to keypoint position changes according to variations in the drone’s pose.
    To achieve this, we analyze the arrangement of the ground-truth drone keypoints by computing the covariance matrix by
    \begin{equation}
    \Sigma = \frac{1}{K} \sum_{k=1}^{K} (\mathbf{y}_{k}^{gt}-\boldsymbol{\mu})(\mathbf{y}_{k}^{gt}-\boldsymbol{\mu})^{T}
    \end{equation}
    where $K$ is the number of keypoints and $\boldsymbol{\mu}$ is the average coordinates. 
    This covariance matrix represents the distribution of drone keypoints as shown in Fig. \textcolor{blue}{\ref{fig:03}}.
    Therefore, we then measure the Mahalanobis distance between the ground-truth keypoints $\mathbf{y}_{k}^{gt}$ and predicted keypoints $\mathbf{y}_{k}^{pred}$ as follows:
    \begin{equation}
        d\left(\mathbf{y}_{k}^{gt},\mathbf{y}_{k}^{pred}\right) = \sqrt{(\mathbf{y}_{k}^{gt}-\mathbf{y}_{k}^{pred})^{T}\Sigma^{-1}(\mathbf{y}_{k}^{gt}-\mathbf{y}_{k}^{pred})}.
    \end{equation}
    Based on this distance metric, we can adaptively measure each keypoint error relative to the drone's pose.
       
    To enhance the stability of the loss function, we design it to operate differently during the early and late stages of training.
    Specifically, in the early stages, when the network is not yet well optimized, the loss function is formulated to be more flexible, allowing for smoother convergence.
    Conversely, in the later stages, the loss function becomes more stringent to enforce precise optimization and improve the model’s final performance.
    To achieve this, we multiply the scale value to the covariance matrix by
    \begin{equation}
    \Sigma_t = S(t)\Sigma+\epsilon \mathbf{I}, \quad \text{where} \quad  S(t)= D e^{-0.01\alpha t},
    \label{eq:09}
    \end{equation}
    where $\alpha$ represents the decay factor, and $D$ denotes the scale factor, and $t$ denotes the training epoch.
    The scale function $S(t)$ exponentially decreases over $t$. 
    To prevent $\Sigma_t$ from becoming a zero matrix, a small identity matrix $(\epsilon \mathbf{I})$ was added.
    Based on this, the covariance matrix $\Sigma_t$ decreases from a large value to a smaller value over the training epoch $t$ as shown in Fig. \textcolor{blue}{\ref{fig:04}}.

    Finally, we reformulate the equation (\textcolor{blue}{8}) with $\Sigma_t$, and the distance function of $t$ is defined as follows:
    \begin{equation}
        d_t\left(\mathbf{y}_{k}^{gt},\mathbf{y}_{k}^{pred}\right) = \sqrt{(\mathbf{y}_{k}^{gt}-\mathbf{y}_{k}^{pred})^{T}\Sigma_t^{-1}(\mathbf{y}_{k}^{gt}-\mathbf{y}_{k}^{pred})}.
    \end{equation}
    The final loss function based on Gaussian distribution is defined as
    \begin{equation}
    \mathcal{L}_{\text{pose}}(t) = \mathbb{E}\left[1- \frac{1}{2\pi|\Sigma_t|^{1/2}}\mathrm{exp}\left(-\frac{1}{2}d_t(\mathbf{y}_{k}^{gt},\mathbf{y}_{k}^{pred})\right)\right].
    \end{equation}
    The proposed loss function adjusts its computation intensity based on the training epoch $t$, as shown in Fig. \textcolor{blue}{\ref{fig:05}}.
    Moreover, the proposed distance $d_t\left(\mathbf{y}_{k}^{gt},\mathbf{y}_{k}^{pred}\right)$ enables adaptive loss computation based on the drone's pose.
    Thus, we designed a loss function that considers both adaptability and stability during training.
    All parameters in our networks are trainable.

%■■■■■■■■■■■■■■■■■■■■■■■■■■■■■■■■■■■■■■■■■■■■■■■■■■■■■■■■■■■■■■■■■■■■■■■■■■■

%■■■■■■■■■■■■■■■■■■■■■■■■■■■■■■■■■■■■■■■■■■■■■■■■■■■■■■■■■■■■■■■■■■■■■■■■■■■
    \subsection{3D Pose Estimator}
\begin{figure}[t]
        % \vspace{-10pt}
    	\centering
    	\includegraphics[width=1\linewidth]{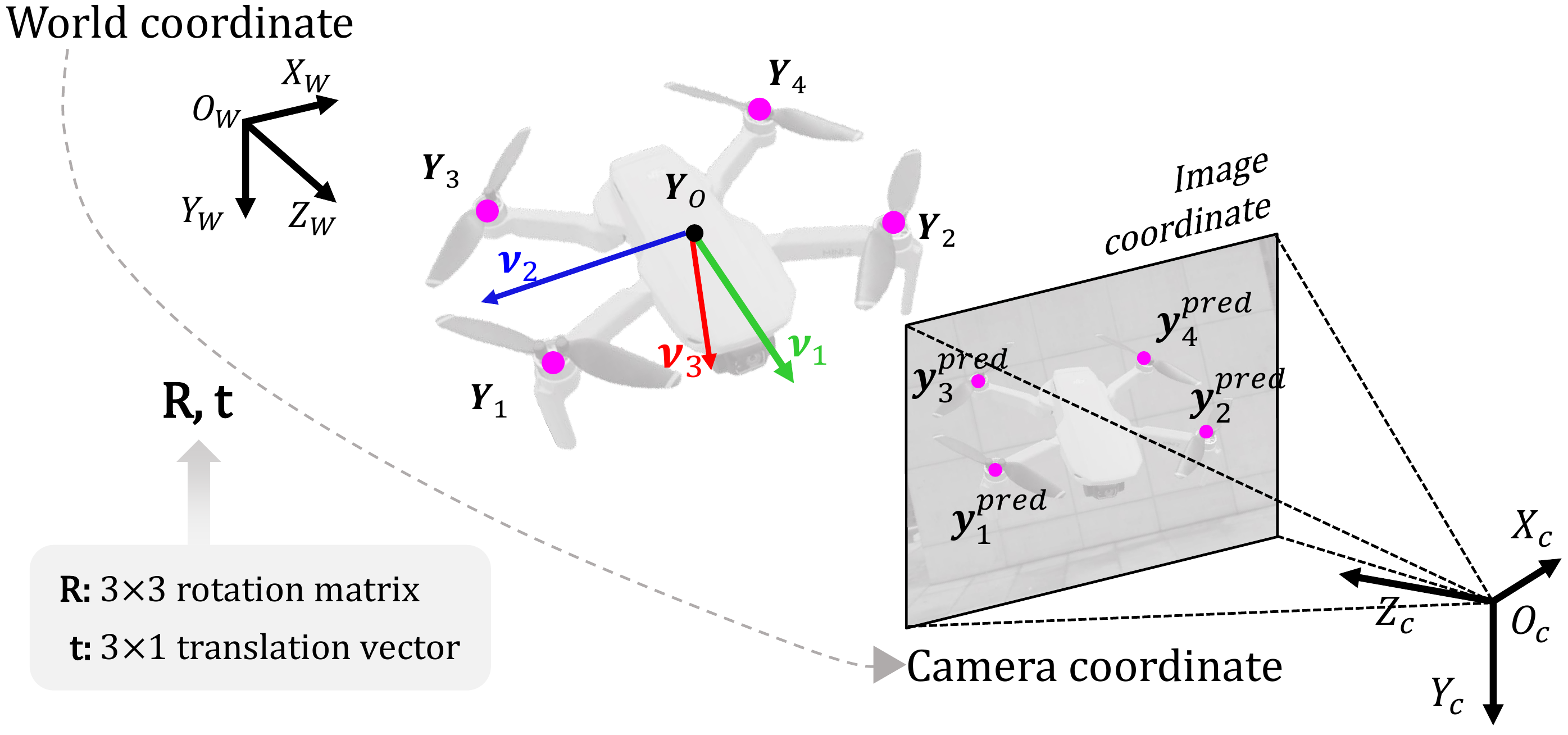}
    	\caption{\textbf{Mapping 2D keypoints of the drone across image, camera, and world coordinate systems for 3D pose (6DoF) estimation.}}
    	\label{fig:06}
        \vspace{-15pt}
    \end{figure}

    The 3D pose of a drone is represented in $6$ degrees of freedom (6DoF). %%%%% ■
    The 6DoF of a drone consists of translation $(t_x, t_y, t_z)$ and rotation angles $(r_x, r_y, r_z)$.
    The rotation can be represented by a $3\times3$rotation matrix $\mathbf{R}_{pose} \in \mathbb{R}^{3\times 3}$,
    and the translation can be represented by a translation vector $\mathbf{t}_{pose} \in \mathbb{R}^{3\times 1}$.
    To estimate the 6DoF using the predicted 2D drone keypoints, the keypoint coordinates are transformed into 3D spatial coordinates, as shown in Fig. \textcolor{blue}{\ref{fig:06}}.
    The image coordinate system contains predicted drone keypoints $\mathbf{y}^{pred} = \{ \mathbf{y}^{pred}_k \}_{k=1}^{K}$, representing detected 2D propeller positions, while the world coordinate system contains corresponding 3D keypoints $\mathbf{Y}^{obj} = \{ \mathbf{Y}^{obj}_k \}_{k=1}^{K}$. 
    Note that, since the type of drone is assumed to be known, the 3D keypoint locations $\mathbf{Y}^{obj}$ of the drone are known.
    For example, an object detector can be utilized to identify the drone class, and by referring to the specifications of the corresponding drone, the actual 3D positions of the keypoints can be obtained.
    In addition, the camera is already calibrated, so camera parameters, such as the intrinsic parameters $\mathbf{A}$, are given.
    
    The transformation from world to camera coordinates is defined by $\mathbf{P}_c = \mathbf{R} \mathbf{X}_w + \mathbf{t}$, where $\mathbf{P}_c = (X_c, Y_c, Z_c)^T$ are camera coordinates.
    The camera pose $(\mathbf{R}, \mathbf{t})$ is estimated using the PnP solver\textcolor{blue}{\cite{pnp}} by minimizing the re-projection error,
    \begin{equation}
    \underset{\mathbf{R}, \mathbf{t}}{\arg\min} \sum_{k=1}^{K} \left| \mathbf{y}^{pred}_k - \mathbf{A}( \mathbf{R} \mathbf{Y}^{obj}_k + \mathbf{t} ) \right|^2,
    \end{equation}
    % ♣ 여기 빠졌음! -> the
    where $\mathbf{A}$ represents the camera intrinsic matrix.
    The 3D keypoint position in camera coordinate is reconstructed by
    \begin{equation}
        \mathbf{Y}_{k} = \mathbf{R}^{-1} (\mathbf{Y}_k^{obj} - \mathbf{t}).
    \end{equation}
    For rotation estimation, the centroid of the four keypoints is computed as $\mathbf{Y}_O = \frac{1}{K} \sum_{k=1}^{K} \mathbf{Y}_k$.
    Using this centroid, two-directional vectors are defined:  
    \(\mathbf{v}_1\) is represented by the green line, and \(\mathbf{v}_2\) is represented by the blue line as in Fig. \textcolor{blue}{\ref{fig:06}}.
    \begin{equation}
    \mathbf{v}_1 = \mathbf{Y}_2 - \mathbf{Y}_O, \quad
    \mathbf{v}_2 = \mathbf{Y}_3 - \mathbf{Y}_O.
    \end{equation}
    The normal vector, represented by the red line, is obtained via the cross product defined by $\mathbf{v}_3 = \mathbf{v}_1 \times \mathbf{v}_2$.
    
    The drone's rotation matrix is formed as:
    \begin{equation}
    \mathbf{R}_{pose} = [\hat{\mathbf{v}}_1 \quad \hat{\mathbf{v}}_2 \quad \hat{\mathbf{v}}_3],
    \end{equation}
    where $\hat{\mathbf{v}}_1$, $\hat{\mathbf{v}}_2$ and $\hat{\mathbf{v}}_3$ are the unit vectors of $\mathbf{v}_1$,  $\mathbf{v}_2$ and 
    % $\mathbf{v}_3$. The estimated rotation $\mathbf{R}_{pose}$ and the centroid position $\mathbf{t}_{pose} =  \mathbf{Y}_O$ form the final 6DoF of the drone.
    $\mathbf{v}_3$. The estimated rotation $\mathbf{R}_{pose}$ and the centroid position $\mathbf{t}_{pose} = \mathbf{Y}_O$ are filtered using a Kalman filter to form the final 6DoF of the drone.
    %By leveraging both the predicted 2D keypoints $\mathbf{y}^{pred}$ and the corresponding 3D keypoints $\mathbf{Y}^{obj}$, the drone’s full 6DoF pose is reconstructed.

%■■■■■■■■■■■■■■■■■■■■■■■■■■■■■■■■■■■■■■■■■■■■■■■■■■■■■■■■■■■■■■■■■■■■■■■■■■■

\section{DATASETS}
\label{sec:datasets}

    All datasets summarized in Tab. \textcolor{blue}{\ref{tab:01}} have been publicly released.
    Since acquiring real-world keypoint and 3D pose ground truth for drones is challenging, we conducted experiments using a synthetic dataset with precise pose information. 
    We provide two new drone datasets: \texttt{2Dronekey} and \texttt{3DronePose}.
    Further details of our datasets are provided in the supplementary material.
    
    \subsection{\texttt{2DroneKey} (2D Images + keypoints)}
    \label{sec:sub:2drone}
    \texttt{2DroneKey} is a dataset containing 2D drone images and propeller keypoints.  
    We used a 3D program to generate a dataset \texttt{2DroneKey} featuring Air2S and Mini2 drone models in five real-world backgrounds captured with a 360-degree camera.  
    It consists of 10 sequences (Seq.1 to Seq.10), each with 1K frames, for a total of 10K frames.
    Rotation varies significantly, while translation remains nearly fixed.  
    Images have a resolution of $1920\times1080$.

    \subsection{\texttt{3DronePose} (6DoF)}
    \label{sec:sub:syn6drone}
    % ♣ 여기 빠졌음! -> additional, actual
    \texttt{3DronePose} provides three sequences (Seq. 11 to Seq. 13) that include 6DoF annotations for drones as well as keypoint annotations.
    Since no public 3D drone 6DoF dataset with keypoints has been available, we created a new dataset by adding 6DoF information to \texttt{2DroneKey}, considering the drone sizes.
    This dataset includes various motion patterns and is designed to evaluate the performance of 3D pose estimation methods.
    Seq.~11 consists of linear motion without rotation, while Seq.~12 exhibits non-linear motion without rotation. In contrast, Seq.~13 involves non-linear motion with rotation.
    All images are captured at a resolution of $1920\times1080$, ensuring high visual quality to enhance the accuracy of keypoint detection and 3D pose estimation.

%■■■■■■■■■■■■■■■■■■■■■■■■■■■■■■■■■■■■■■■■■■■■■■■■■■■■■■■■■■■■■■■■■■■■■■■■■■■

        \begin{table*}[t]
        % \vspace{5pt}
            \centering
            \setlength{\tabcolsep}{5pt} % Adjust column spacing
            \renewcommand{\arraystretch}{1} % Adjust row spacing
                \begin{minipage}{0.48\linewidth}
                \centering
                \caption{\textbf{Performance by number of layers.}}
                \label{tab:04}
                \begin{tabular}{c|cc|c|c}
                    \hline\noalign{\hrule height 0.5pt}
                    \textbf{Layer Size} & $SR^{kp}_{90}$ & $SR^{kp}_{95}$& $AP^{kp}$ & Number of parameters \\ 
                    \hline\noalign{\hrule height 0.5pt}
                    2                   & 98.3  & 95.3  & 99.1  & 42.41M  \\ 
                    3                   & 99.1  & 96.5  & 99.4  & 59.21M  \\ 
                    \textbf{6}          & \textbf{99.8} & \textbf{98.5} & \textbf{99.7} & 84.41M  \\ 
                    8                   & 99.7  & 97.8  & 99.6  & 101.22M \\ 
                    10                  & 99.4  & 97.8  & 99.6  & 118.02M \\ 
                    12                  & 98.2  & 94.9  & 99.14 & 134.82M \\ 
                    \hline\noalign{\hrule height 0.5pt}
                \end{tabular}
                \\ \vspace{2pt} * Feature dimension: 1024; loss function: $\mathcal{L}_{pose} (\alpha = 5, D = 10)$ 
            \end{minipage}
            \hfill
            \begin{minipage}{0.48\linewidth}
                \centering
                \caption{\textbf{Performance by embedding $d$.}}
                \label{tab:03}
                \begin{tabular}{c|cc|c|c}
                    \hline\noalign{\hrule height 0.5pt}
                    \textbf{Dimension} & $SR^{kp}_{90}$ & $SR^{kp}_{95}$ & $AP^{kp}$ & Number of parameters \\ 
                    \hline\noalign{\hrule height 0.5pt}
                    128       & 98.1  & 93.6  & 99.1  & 27.92M \\ 
                    256       & 98.7  & 96.7  & 99.4  & 33.24M \\ 
                    512       & 97.3  & 95.5  & 99.1  & 46.63M \\ 
                    768       & 99.7  & 96.8  & 99.5  & 63.69M \\
                    \textbf{1024} & \textbf{99.8} & \textbf{98.5} & \textbf{99.7} & 84.42M \\ 
                    2048      & 98.7  & 95.8  & 99.4  & 199.84M \\ 
                    \hline\noalign{\hrule height 0.5pt}
                \end{tabular}
                \\ \vspace{2pt} * Number of layers: 6; loss function: $\mathcal{L}_{pose} (\alpha = 5, D = 10)$ 
            \end{minipage}
            \vspace{-5pt}
        \end{table*}

        \begin{figure*}[t]
	\centering
        % \subfigure[Layer size = 2] {\includegraphics[width=0.725\columnwidth]{figures/figure_gate/05_2_gate_output.png}}\hspace{-20pt}
		\subfigure[Layer size = 3] {\includegraphics[width=0.5\columnwidth]{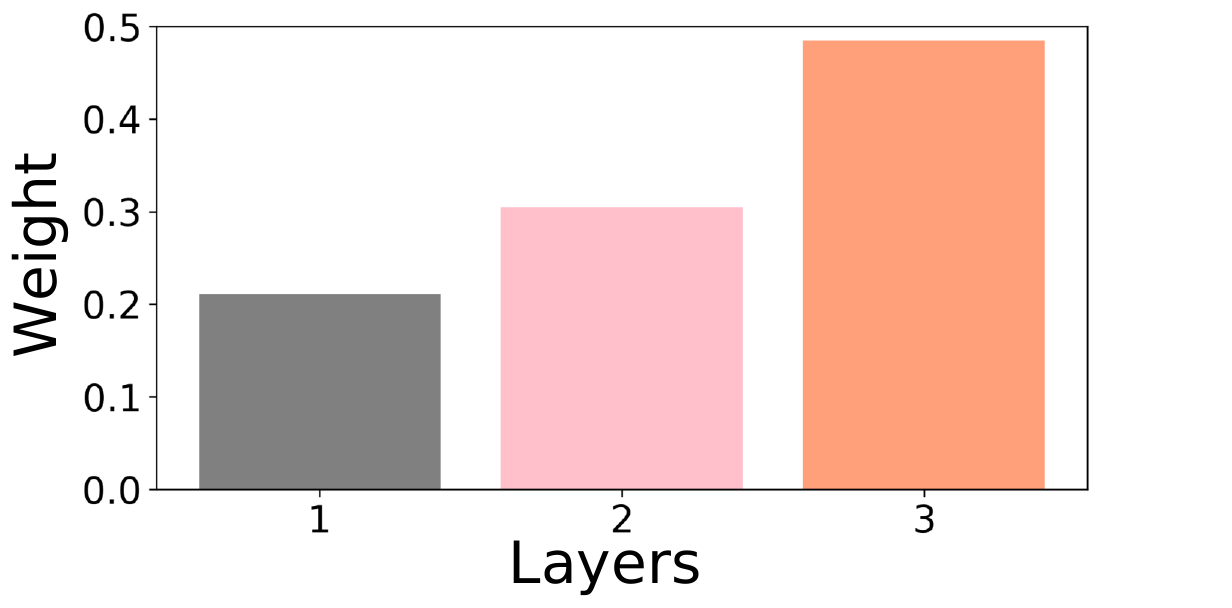}}
		\subfigure[Layer size = 6] {\includegraphics[width=0.5\columnwidth]{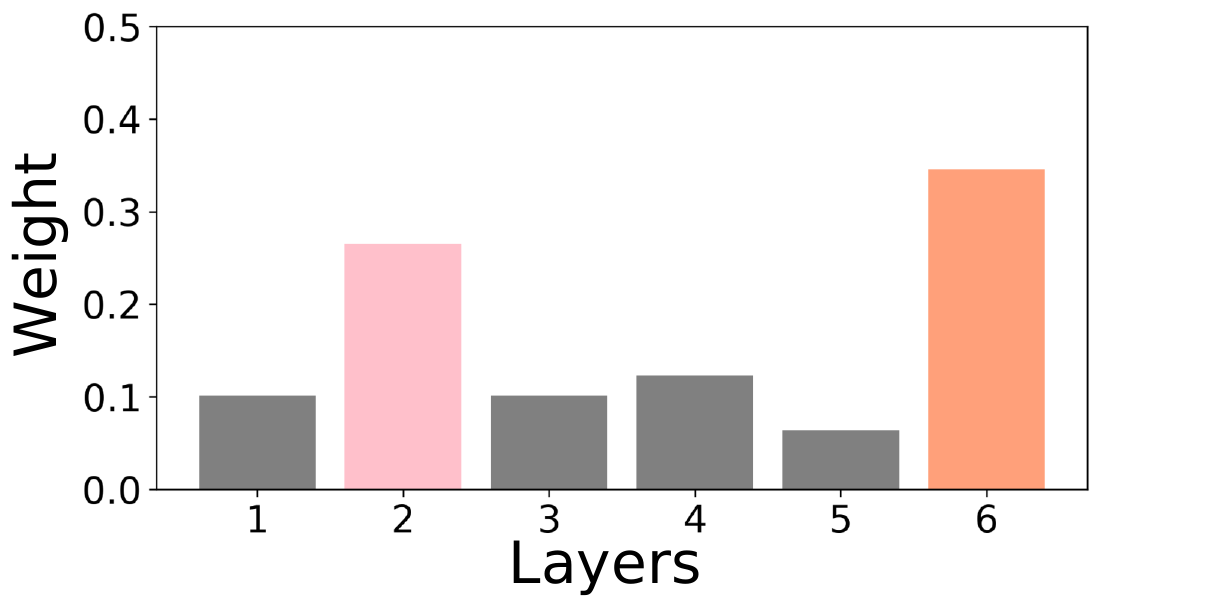}}
            \subfigure[Layer size = 8] {\includegraphics[width=0.5\columnwidth]{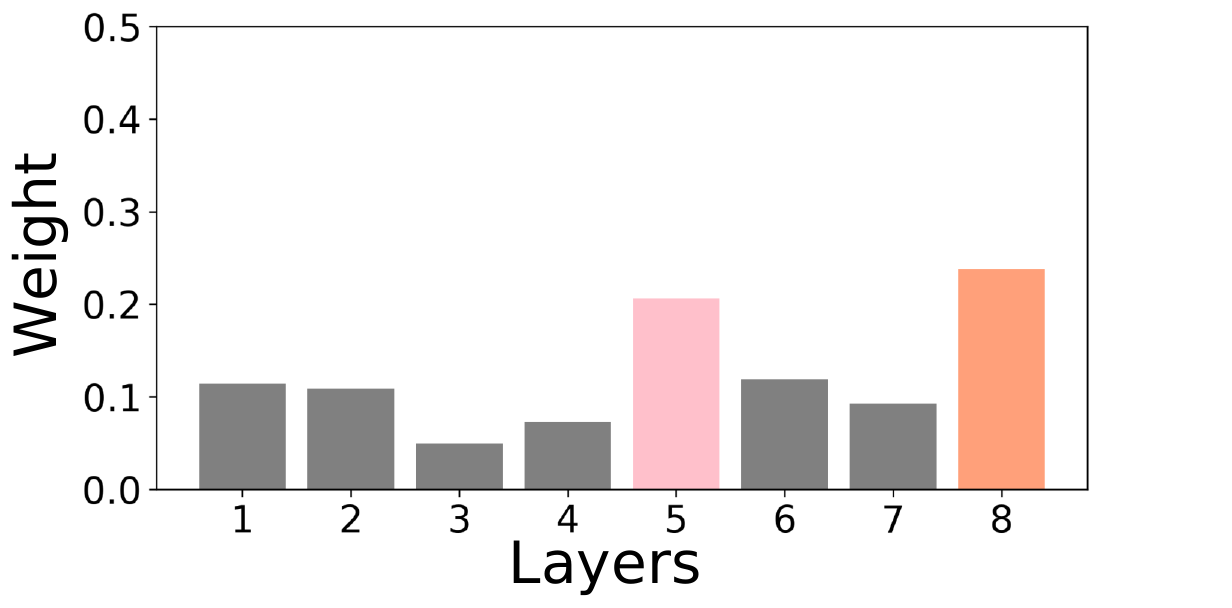}} 
            % \subfigure[Layer size = 10]{\includegraphics[width=0.725\columnwidth]{figures/figure_gate/05_10_gate_output.png}}\hspace{-20pt}
            \subfigure[Layer size = 12]{\includegraphics[width=0.5\columnwidth]{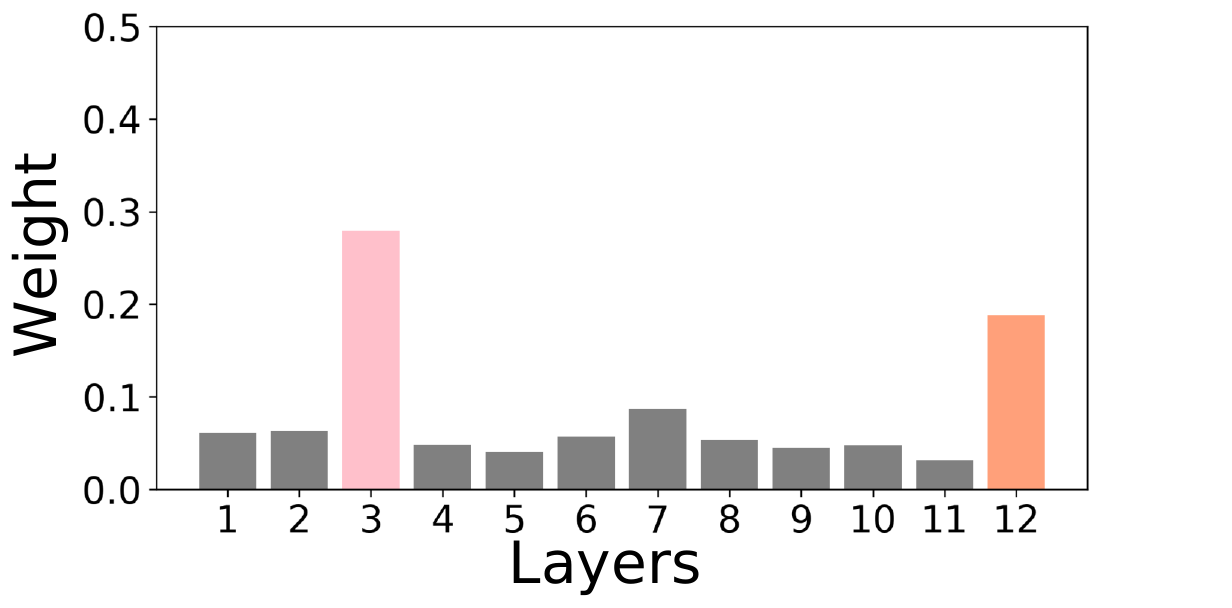}} \vspace{-10pt}
            \caption{\textbf{Gate weights analysis for different layer numbers.}
            \textcolor{pink}{■} represents the highest weight values in intermediate layers (excluding the final layer), and \textcolor{orange}{■} the final layer weights. This analysis highlights the significant contribution of both final and intermediate layers to keypoint learning, emphasizing the importance of features extracted at different network stages.}
	\label{fig:08}\vspace{-10pt}
        \end{figure*}

%■■■■■■■■■■■■■■■■■■■■■■■■■■■■■■■■■■■■■■■■■■■■■■■■■■■■■■■■■■■■■■■■■■■■■■■■■■■

\section{EXPERIMENTAL RESULTS}
\label{sec:experimental_results}

    \subsection{Settings and Evaluation Metrics}
    
    % \subsubsection{Calibration}
    % 굳이...? -> 한 줄로라도 짧게 넣을 것 ★★★★★★★★★
    % Accurate 3D reconstruction requires camera calibration. In this study, offline calibration was used to ensure spatial accuracy, as drones are often in environments with monotonous backgrounds, making online calibration difficult.
    
    % Ubuntu with an NVIDIA RTX 3060 GPU.
    All experiments were conducted on an Ubuntu system with an NVIDIA RTX 3060 GPU.
    % The model -> transformer: encoder layer size 6, head 8, dimension 1024
    Our model is based on a transformer encoder with 6 layers, 8 attention heads, and a feature dimension of 1024.
    % training setting -> 100 epoch, batch size of 8, a leraning rate of $10^{-5}$, adap optimizer
    The model was trained for 100 epochs with a batch size of 8, a learning rate of $10^{-5}$, and the Adam optimizer.
    % for pose-adaptive 마할라비스 loss -> \alpha 5, D = 10
    For the pose-adaptive Mahalanobis loss function, we set $\alpha=5$ and $D = 10$.
    % ablation study를 통해 세팅된 최적의 값임 (sec~~) ★★★★★★★★★
    These settings were determined as the optimal values through the ablation study.

    In this study, object keypoint similarity (OKS)\textcolor{blue}{\cite{oks}} is used to evaluate keypoint detection.
    OKS measures the similarity between predicted and ground truth (GT) keypoints on a scale from $0$ to $1$, where $1$ indicates perfect alignment and 0 indicates no similarity. It is defined as:
    \begin{equation}
    \small
        OKS_{drone} = \frac{1}{K} \sum_{k=1}^{K} \exp\left(-\frac{d_k^2}{0.2\beta^2\gamma_k^2}\right)
    \end{equation}
    where $K$ is the number of keypoints and $d_k$ is the Euclidean distance between predicted ($\mathbf{y}^{pred}$) and ground truth ($\mathbf{y}^{GT}$) keypoints. $\beta$ represents the square root of the object area. Drones require more precise detection, so a stricter evaluation with a value of 0.2 is used. $\gamma$ is the keypoint weight, we set to the same weight 1 for all drone propellers.
    To quantitatively evaluate keypoint detection, we used three metrics based on OKS:
    \begin{itemize}
        \item $SR^{kp}_{90}$: success rate at an OKS threshold of $0.90$ \\ (percentage of keypoints with $OKS \geq 0.90$).
        \item $SR^{kp}_{95}$: success rate at an OKS threshold of $0.95$ \\ (percentage of keypoints with $OKS \geq 0.95$).
        \item $AP^{kp}$: average precision (AP), computed as the mean OKS-based precision over multiple thresholds.
    \end{itemize}
    
    3D drone pose is represented in 6DoF, including rotation and translation.
    % Rotation(회전) 오차는 예측된 자세와 실제 자세 간의 각도 차이를 angle space에서 계산하여 평가한다. 
    Rotation error is computed as the angle difference between the predicted and GT orientations.
    % 모든 테스트 시퀀스에서 각 프레임별 오차를 구한 후, 시퀀스 전체에 대해 평균 각도 오차(MAE)를 계산한다.
    % 값이 낮을수록 예측 정확도가 높다.
    The mean angle error (MAE-angle) is averaged over all test sequences, where lower values indicate better accuracy.
    % 
    % Translation(이동) 오차는 예측된 위치와 실제 위치 간의 유클리드 거리 차이를 계산하여 평가한다. 
    Translation error is measured as the Euclidean distance between predicted and GT positions.
    % 모든 테스트 시퀀스에서 각 프레임별 오차를 구한 후, 시퀀스 전체에 대해 평균을 내어 MAE-absolute 또는 RMSE 지표로 성능을 측정한다. 
    % 값이 낮을수록 예측 정확도가 높다.
    It is evaluated using root mean square error (RMSE) and mean absolute error (MAE-absolute), with lower values indicating higher accuracy.
    
%■■■■■■■■■■■■■■■■■■■■■■■■■■■■■■■■■■■■■■■■■■■■■■■■■■■■■■■■■■■■■■■■■■■■■■■■■■■
    \begin{table}[t]
        \centering
        % \vspace{-15pt}
        \caption{\textbf{Performance comparison of loss functions.} 
                This table compares the performance of different loss functions for keypoint detection, including MSE loss, Gaussian loss, and pose-adaptive Mahalanobis loss.}
        \label{tab:05}
        % \small
        \footnotesize
        {\renewcommand{\arraystretch}{1} % 이 테이블에서만 적용
        \begin{tabular}{c|c|cc|c}
        \hline\noalign{\hrule height 0.5pt}
            \multirow{1}{*}{\textbf{Loss Type}} & \multirow{1}{*}{Decay Type} & \multirow{1}{*}{$SR^{kp}_{90}$} & \multirow{1}{*}{$SR^{kp}_{95}$} & \multirow{1}{*}{$AP^{kp}$} \\ 
            &  & & &  \\ \hline \noalign{\hrule height 0.5pt}
            MSE Loss                   & -                                   & 99.1            & 96.9           & 99.5\\ \hline
            \multirow{2}{*}{\makecell{Gaussian\\ Loss}}     
                                           & Fixed                           & 97.8            & 94             & 98.9\\ 
                                           % & Fixed                           & 99.6            & 97.4           & 99.6\\ 
                                           & Exp ($\alpha$=5, $D$=10)  & 99.4            & 97.4           & 99.5\\ \hline
            \multirow{4}{*}{\makecell{Pose-adaptive\\ Loss}}  
                                           & Fixed                           & 99.3            & 97.5           & 99.6\\  
                                           % & Fixed                           & 93.3            & 89             & 96.3\\  
                                           & Linear                   & 97.1            & 95.4           & 98.9\\  
                                           & Exp ($\alpha$=5,  $D$=10)  & \textbf{99.8}   & \textbf{98.5}  & \textbf{99.7} \\
                                           & Exp ($\alpha$=10, $D$=10) & 92.1            & 88             & 95.3\\  \hline  
            \noalign{\hrule height 0.5pt}
        \end{tabular}
        } \\ \vspace{2pt} * Number of layers: $6$; feature dimension: $1024$ \vspace{-5pt}
    \end{table}

    \begin{figure}[t]
       \centering
        \includegraphics[width=0.9\linewidth]{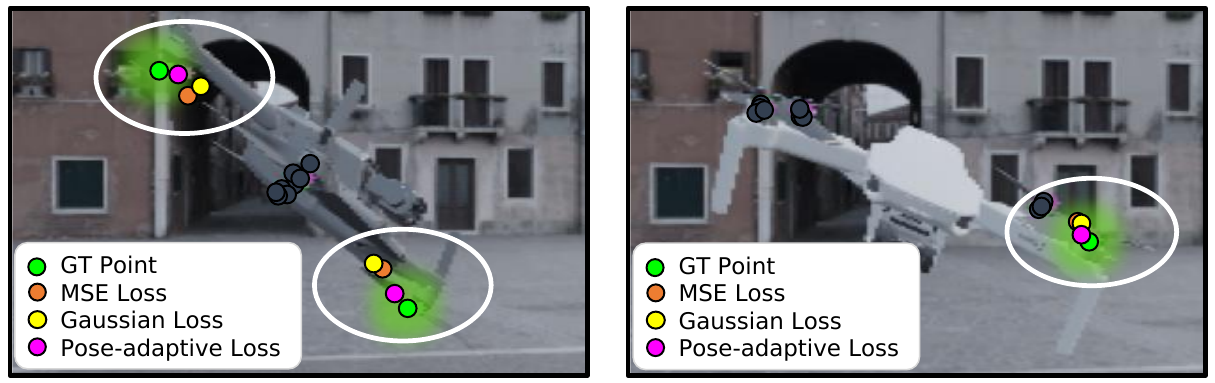}
    \caption{\textbf{Visualization of keypoint detection with different loss functions.} Pose-adaptive Mahalanobis loss \textcolor{magenta}{●} predicts keypoints closer to the GT \textcolor{green}{●}, than MSE and Gaussian loss.}
        \label{fig:09} \vspace{-15pt}
    \end{figure}

    \begin{table*}[t]
    % \vspace{5pt}
        \centering
        \caption{\textbf{Performance comparison of keypoint detection.}
        This table presents a comparative analysis of different methods for keypoint detection, categorized into object detectors, heatmap-based keypoint detectors, and coordinate-based keypoint detectors. ( * 6 encoder layers + 6 decoder layers = 12 layers)}
        \label{tab:02}
        \vspace{-5pt}
        \resizebox{\textwidth}{!}{
        % \footnotesize
        \scriptsize
        {\renewcommand{\arraystretch}{1} % 이 테이블에서만 적용
        \setlength{\tabcolsep}{10pt}
            \begin{tabular}{cc|c|c|cc|c}
            \hline\noalign{\hrule height 0.5pt}
                \multicolumn{2}{c|}{Approaches \& Methods}                                                               & Baseline & Layer                   & $SR^{kp}_{90}$ & $SR^{kp}_{95}$ & $AP^{kp}$ \\ \hline\noalign{\hrule height 0.5pt}
                \multicolumn{1}{c|}{\multirow{2}{*}{Object detector}}     & YOLOv8\textcolor{blue}{\cite{yolov8}} & CNN       & -                   & 69.6           & 69.6            & 70.8       \\ 
                \multicolumn{1}{c|}{}                                     & DETR\textcolor{blue}{\cite{detr}}      & Transformer     & 12*        & 81.4           & 42.2            &  80.9          \\ \hline
                \multicolumn{1}{c|}{\multirow{2}{*}{Heatmap-based keypoint detector}} & ViTPose\textcolor{blue}{\cite{vitpose}}  & Transformer    & 12           & 9.8            & 6.2             & 21.32          \\
                \multicolumn{1}{c|}{}                                  & TokenPose\textcolor{blue}{\cite{tokenpose}} & Transformer & 12 & 32.5           & 31.4            & 43.6      \\ \hline
                \multicolumn{1}{c|}{{\multirow{2}{*}{Cooridnate-based keypoint detector}}}  & SimCC\textcolor{blue}{\cite{simcc}} & CNN & -        & 37.3           & 34.3            & 59.1           \\ \cline{2-7} 
                \multicolumn{1}{c|}{}                                     & \textbf{Ours}& \textbf{Transformer} & \textbf{6}      &\textbf{99.8}   &\textbf{98.5}    &\textbf{99.68} \\ \hline\noalign{\hrule height 0.5pt}
            \end{tabular}
        }}
           \vspace{-5pt}
    \end{table*}

    \begin{figure*}[t]
	\centering
            \subfigure[Case 1] {\includegraphics[width=0.55\columnwidth]{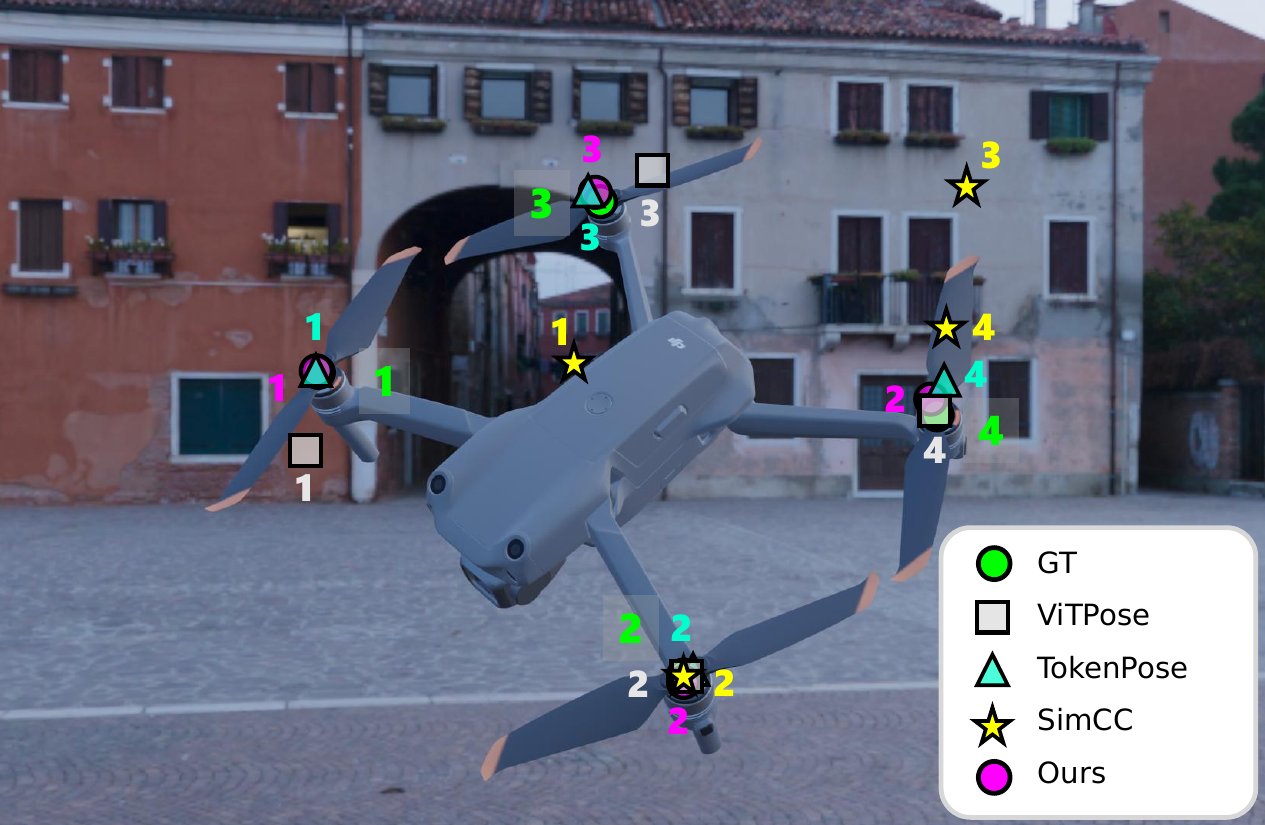}}\hspace{30pt}
		% \subfigure[Case 2] {\includegraphics[width=0.6\columnwidth]{figures/figure_key/figure_key_02.pdf}}\hspace{20pt}
		% \subfigure[Case 3] {\includegraphics[width=0.6\columnwidth]{figures/figure_key/figure_key_03.pdf}}
  %           \subfigure[Case 4] {\includegraphics[width=0.6\columnwidth]{figures/figure_key/figure_key_04.pdf}}\hspace{20pt}
            \subfigure[Case 5] {\includegraphics[width=0.55\columnwidth]{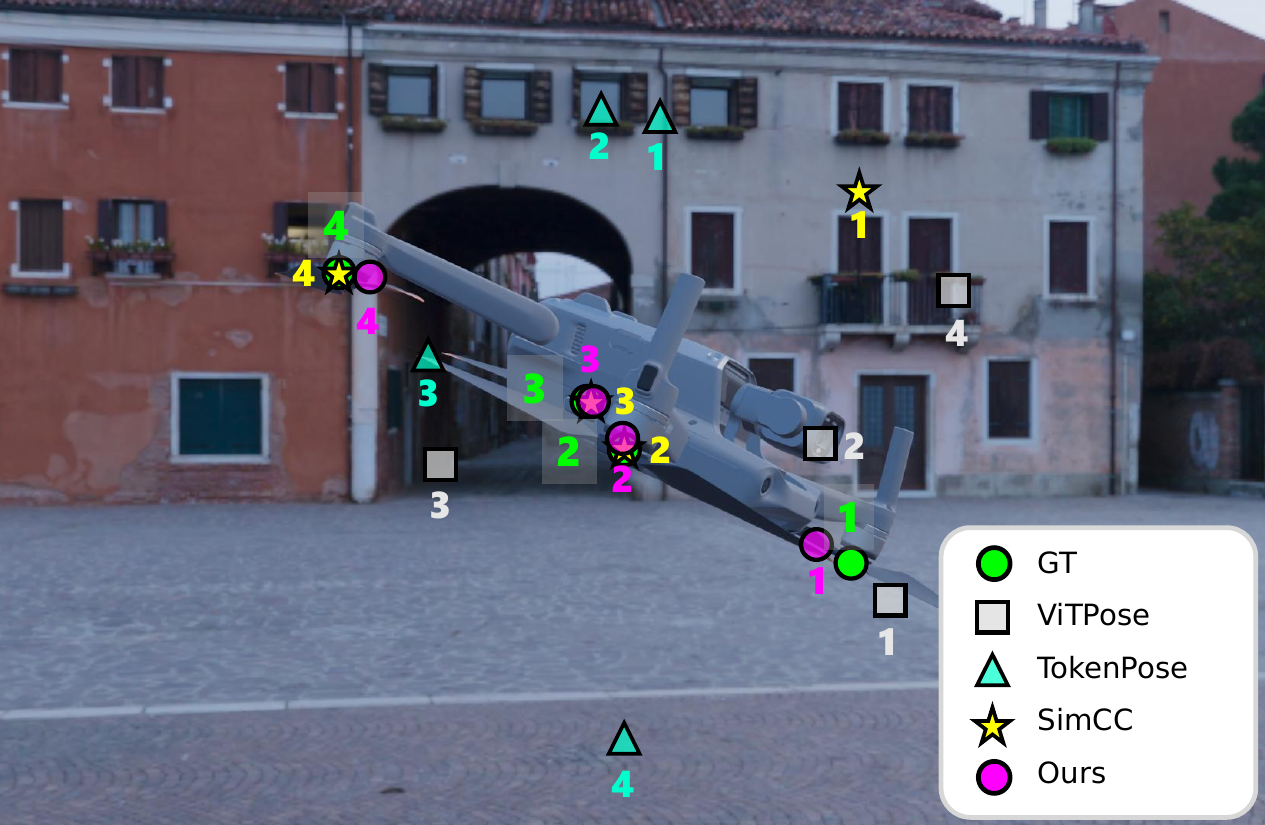}}\hspace{30pt}
            \subfigure[Case 6] {\includegraphics[width=0.55\columnwidth]{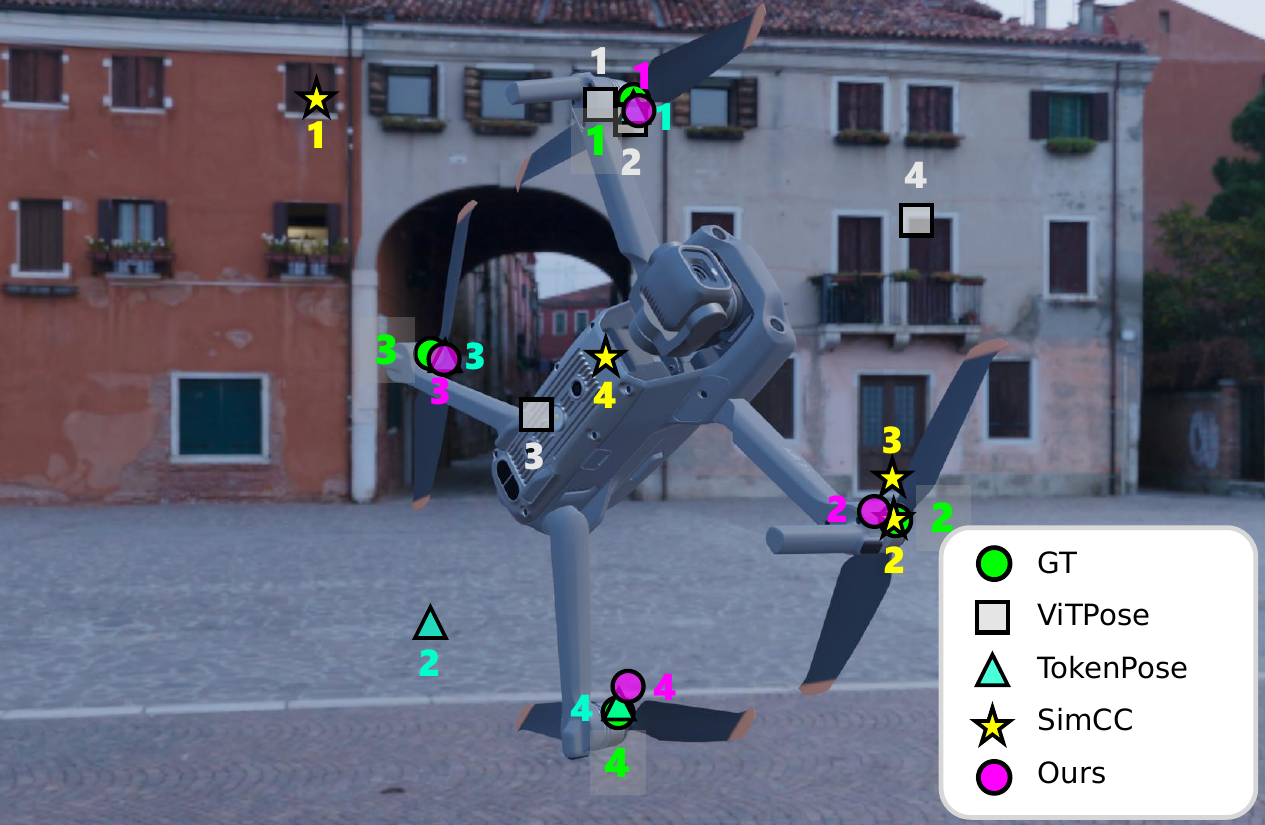}}\vspace{-7pt}
	\caption{\textbf{keypoint detection results for comparative experiments.}
                The predictions of ViTPose\textcolor{blue}{\cite{vitpose}}, TokenPose\textcolor{blue}{\cite{tokenpose}}, SimCC\textcolor{blue}{\cite{simcc}} and the proposed method DroneKey (ours) are visualized alongside the GT in each case. 
                The numbers ($1$ to $4$) indicate the index of the keypoints. Further details of experimental results are provided in
the supplementary material.} \vspace{-15pt}
	\label{fig:07}
    \end{figure*}

    \subsection{Ablation Studies of Keypoint Detection}
    %%% Masked autoencoder are scalable vision learners paper에서는 ablation study부터 소개했었음 -> 논리 구성이 이게 맞는 것 같음...
    % 앞서 설명한 settings의 model이나 loss 파라미터 등을 ablation study를 통해 확립함

    \subsubsection{\textbf{Encoder layers}}
    %% ■ Encoder layer 개수 -> 먼저 Encoder layer의 개수를 ablate함. heatmap regression 은 12개의 encoder를 사용하는데, 우리 방법은 6개에서 가장 좋은 성능을 보였음...
    % tab.2에서 볼 수 있듯 우리는 transformer의 encoder layer 개수를 ablate했다.
    As shown in Tab. \textcolor{blue}{\ref{tab:04}}, we ablated the number of encoder layers.
    % 이때 임베딩 차원은 d = 1024로, loss function은 우리가 제안한 pose-adaptive 마할라노비스 loss를 사용했다.
    In all evaluations, the embedding dimension was set to $1024$, and the proposed loss function $\mathcal{L}_{pose}$ was used.
    % 결과적으로, 6개 layer 구조는 모든 지표에서 가장 높은 성능을 제공하면서도, 8개 이상의 layer 구조보다 적은 파라미터를 유지하여 가장 균형 잡힌 형태를 보였다.
    The results show that the 6-layer structure achieved the best performance while maintaining a well-balanced number of parameters.
    % 반면, layer 개수가 2 또는 3개로 줄어들면 적은 파라미터를 사용하지만, AP 성능이 6개 layer보다 낮았다.
    In contrast, models with 2 and 3-layer had fewer parameters but lower performance than the 6-layer.
    % 8개와 10개 layer에서는 6개 layer보다 AP와 SR이 낮고, 파라미터 수도 증가했다.
    % 또한, For 12 layers, 성능이 현저히 감소했다.
    In the 8, 10, and 12-layer, performance decreased while the number of parameters increased.
    %
    % 우리는 각 encoder layer에서 compact representation을 추출하고, layer별 중요도를 반영하는 gate를 생성하여 최종적으로 sum하는 방식을 적용했다.
    We extracted compact representations from each encoder layer and applied a gated sum to reflect layer importance.
    % Fig.~\textcolor{blue}{\ref{fig:08}}에서 볼 수 있듯, 최종 layer(\textcolor{orange}{■})의 gate 값이 높게 나타났으며, 이는 객체 수준의 정보가 반영되었기 때문으로 해석된다.
    As shown in Fig.~\textcolor{blue}{\ref{fig:08}}, the final layer (\textcolor{orange}{■}) had the highest gate value, reflecting the integration of object-level information.
    % 그러나 중간 layer(\textcolor{pink}{■})에서도 높은 weight가 관찰되었으며, 특히 layer 개수가 6 또는 8일 때 최종 layer와 유사한 수준을 보였다.
    However, the mid-layers (\textcolor{pink}{■}) also showed high weights, especially in 6 and 8-layer models, where they were comparable to the final layer. 
    % 또한, layer가 12개일 경우 오히려 중간 layer의 weight가 더 높아졌다.
    Notably, in the 12-layer, mid-layer weights surpassed those of the final layer. 
    % 이는 key-point 정보가 중간 layer에서 효과적으로 학습되며, layer별 gated sum이 중요한 역할을 함을 보여준다.
    This suggests that keypoint information is effectively learned in mid-layers, and gated summation plays a crucial role in balancing information across layers.

    %% ■ Feature Dimension
    \subsubsection{\textbf{Embedding dimension}}
    % 우리는 이어서 embedding dimension에 대한 ablation study를 진행했다.
    We conducted an ablation study on the embedding dimension.
    % 이때 encoder layer size는 6으로, loss function은 우리가 제안한 pose-adaptive 마할라노비스 loss를 사용했다.
    The encoder had $6$-layer, and the proposed $\mathcal{L}_{pose}$ was used.
    % Tab. \textcolor{blue}{3}은 1024 차원에서 가장 좋은 성능이 달성되며, 이 크기에서 성능이 안정적으로 유지됨을 보여준다. 
    Tab. \textcolor{blue}{\ref{tab:03}} shows that the best performance is achieved at $1024$ dimensions, with stable results at this size. 
    % 2048 차원에서는 파라미터 수가 크게 증가하면서 성능이 감소한다. 
    Performance declines at $2048$ dimensions, where the parameter count increases significantly. 
    % 512 차원 이하에서는 성능이 낮아져, 1024 차원이 정확도와 계산 효율성의 균형을 잘 맞추는 최적의 선택임을 알 수 있다.
    Lower dimensions (below $512$) show reduced performance, making $1024$ dimensions the optimal choice for balancing accuracy and computational efficiency.

    %% ■ Loss
    \subsubsection{\textbf{Loss function}}
    % Tab.~\textcolor{blue}{\ref{tab:05}}에서는 다양한 손실 함수가 키포인트 검출 성능에 미치는 영향을 비교하였다. 
    Tab. \textcolor{blue}{\ref{tab:05}} compares the impact of different loss functions on keypoint detection performance.
    % 실험에 사용된 손실 함수는 MSE Loss, Gaussian Loss, 그리고 우리가 제안한 Pose-adaptive Mahalanobis Loss이다.
    The evaluated loss functions include MSE loss, Gaussian loss, and the proposed pose-adaptive Mahalanobis loss with various decay types: Fixed, Linear, Exp.
    For the exponential decay, we used the proposed scale function $S(t)$ as in Eq~\ref{eq:09}.
    % %
    % Gaussian Loss와 Pose-adaptive Loss는 학습이 진행될수록 더욱 정밀한 예측이 가능하도록 decay factor($\alpha$)와 scale factor($\beta$)를 적용하여 학습 초반에는 유연하게, 후반부에는 MSE Loss에 가까운 형태로 학습되도록 설계하였다.
    Gaussian loss and pose-adaptive Mahalanobis loss incorporate decay factor ($\alpha$) and scale factor ($D$) to improve prediction accuracy as training progresses.
    The proposed exponential decay enables flexibility in the early training stages for smoother convergence and becomes stricter later to enforce precise optimization and enhance final performance.
    
    % Gaussian Loss의 경우, Fixed 설정에서는 MSE Loss보다 낮은 성능을 보였으나, Exponential Decay($\alpha=5, D=10$)를 적용하면 GT에 가까운 키포인트 검출 성능이 향상되었다. 
    Gaussian loss does not estimate covariance $\Sigma$ based on the keypoint distribution but instead uses an identity matrix.
    For Gaussian loss, the fixed setting resulted in lower performance than MSE loss, but applying exponential decay ($\alpha=5, D=10$) improved keypoint detection performances.
    % Pose-adaptive Loss에서는 Exp($\alpha=5, D=10$)설정이 가장 우수한 결과를 보였으며, $AP^{kp}$ 99.7\%, $SR^{kp}{90}$ 99.8\%, $SR^{kp}{95}$ 98.5\%를 기록하였다. 
    The pose-adaptive loss with exponential decay ($\alpha=5, D=10$) achieved the best results, with $AP^{kp}$ of 99.7\%, $SR^{kp}_{90}$ of 99.8\%, and $SR^{kp}_{95}$ of 98.5\%. 
    % 반면, $\alpha$ 값을 증가시키면 학습 초반에 값이 급격히 감소하여 성능이 저하되는 현상이 나타났다.
    In contrast, increasing $\alpha=10$ led to rapid decay in the early training stages, negatively affecting performance.
    % %
    % 결과적으로, Exp($\alpha=5, \beta=10$) 설정을 적용한 Pose-adaptive Loss는 극단적인 드론 포즈에서도 안정적인 성능을 유지하며, 기존 MSE Loss와 Gaussian Loss에서 발생했던 키포인트 오류를 효과적으로 해결하였다(Fig.~\textcolor{blue}{\ref{fig:09}}). 
    Overall, the pose-adaptive Mahalanobis loss with exponential decay ($\alpha=5, D=10$) demonstrated robust performance even in extreme drone poses.
    It effectively addresses keypoint errors observed with MSE and Gaussian loss (Fig. \textcolor{blue}{\ref{fig:09}}). 
    % 이는 학습 과정에서 손실 함수를 동적으로 조정하는 방식이 복잡한 포즈에서도 효과적인 검출을 보장함을 확인시켜준다.
    These findings confirm that dynamically adjusting the loss function during training enhances keypoint detection for complex poses.

%■■■■■■■■■■■■■■■■■■■■■■■■■■■■■■■■■■■■■■■■■■■■■■■■■■■■■■■■■■■■■■■■■■■■■■■■■■■
    \begin{figure*}[t]
    % \vspace{-5pt}
	\centering
		\subfigure[Sequence 11]{\includegraphics[width=0.55\columnwidth]{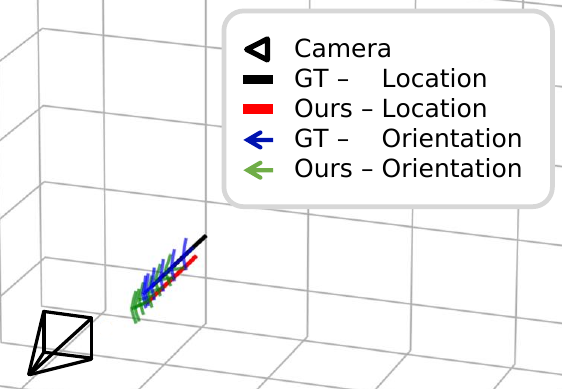}}\hspace{30pt}
		\subfigure[Sequence 12]{\includegraphics[width=0.55\columnwidth]{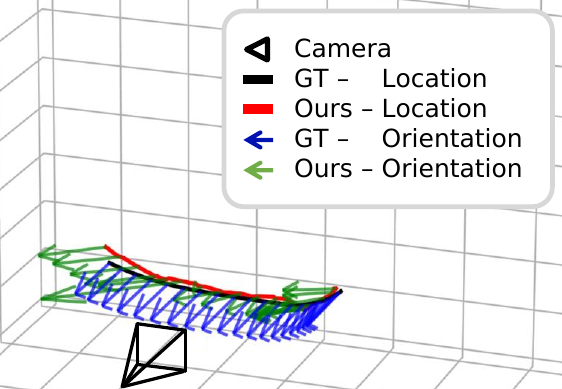}}\hspace{30pt}
		\subfigure[Sequence 13]{\includegraphics[width=0.55\columnwidth]{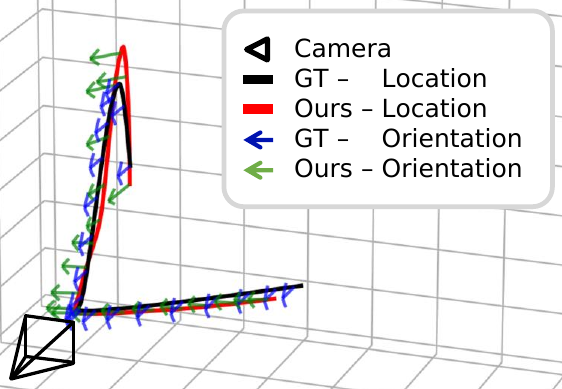}}\vspace{-5pt}
	\caption{\textbf{Qualitative results of the drone 3D pose (6DoF) estimation based on the proposed method.}} \vspace{-15pt}
	\label{fig:10}
        \end{figure*}

    \subsection{Performance Comparison of Keypoint Detection}
    % 내용 다시 확인 필요
    % 앞서 가장 높은 AP를 보인 모델의 setting?을 사용해서 다른 연구와 비교함
    % ♣ 여기 빠졌음! -> including, both, correctly
        Tab. \textcolor{blue}{\ref{tab:02}} compares keypoint detection methods: object detectors, heatmap-based, and coordinate-based approaches.
        The model was trained on 7K images, validated on 2K, and tested on 1K for consistency.
        Keypoint detection requires accurate detection (i.e., location ($x,y$) and size ($w,h$)) and correct ordering (i.e., keypoints labeled as 1, 2, 3, and 4).
        For drone propeller detection, we evaluated YOLOv8\textcolor{blue}{\cite{yolov8}} and DETR\textcolor{blue}{\cite{detr}}, assigning detector labels to encode the order.
        Both methods detected the location and size of the propellers but failed to predict their order due to the high visual similarity among them.
        DETR, leveraging global features, achieved an $AP^{kp}$ of 80.9\%, outperforming YOLOv8.
        
        We also tested TokenPose\textcolor{blue}{\cite{tokenpose}} and ViTPose\textcolor{blue}{\cite{vitpose}}, which employ the heatmap regression approach.
        % As depicted in Fig. \textcolor{blue}{\ref{fig:07}}, they struggled with accruate keypoint detection.
        As depicted in Fig. \textcolor{blue}{\ref{fig:07}}, they struggled with accurate keypoint detection, either detecting incorrect locations (a-c) or confusing the order (c). 
        Moreover, background clutter led to missed detections in both models.
        % In case 6, SimCC mixed up propellers 2 and 3, while ViTPose misidentified propellers 1 and 2.
        % Transformers typically require large-scale training datasets of over 300M images \textcolor{blue}{\cite{300m+}}, but with only 7K images in the drone case, the AP remained low.
        % For coordinate-based methods, SimCC \textcolor{blue}{\cite{simcc}} outperformed TokenPose but still struggled due to the limited training dataset.
        Transformers typically require large-scale training datasets of 300M+ images\textcolor{blue}{\cite{300m+}}, and existing keypoint detection models are commonly trained on human COCO (56K+)\textcolor{blue}{\cite{coco}}.
        Additionally, we set the OKS $\beta$ value to $0.2$ to account for the small keypoint regions of drone propellers.
        As a result, on the \texttt{2DroneKey} (7K), TokenPose, ViTPose, and SimCC recorded low $AP^{kp}$ due to the limited data and stricter evaluation criteria.
        %
        % Additionally, background clutter led to missed detections in both models.
        The proposed DroneKey achieved $AP^{kp}$ 99.68\%, outperforming all previous methods.
        As shown in Fig. \textcolor{blue}{\ref{fig:07}}, it remains highly accurate, even in complex backgrounds and extreme poses.
        This experiment confirms that the proposed DroneKey efficiently learns to accurately detect drone keypoints even with a limited dataset.

    \subsection{Performance of 3D Pose Estimation}

    \begin{table}[t]
    \vspace{-5pt}
    \centering
        \caption{\textbf{Performance of 3D pose (6DoF).}}
        \label{tab:06}
            % \resizebox{\textwidth}{!}{
    \footnotesize
    \begin{tabular}{c|c||ccc|c}
    \hline \noalign{\hrule height 0.5pt}
    \multirow{2}{*}{6DoF}       & \multirow{2}{*}{Metrics} & \multicolumn{3}{c|}{Sequences (\textsc{3DronePose})}                           & \multirow{2}{*}{Avg} \\ \cline{3-5}
                                   &                          & \multicolumn{1}{c|}{Seq. 11} & \multicolumn{1}{c|}{Seq. 12} & Seq. 13          &                         \\ \hline \noalign{\hrule height 0.5pt} 
    $\mathbf{R}_{pose}$               & \footnotesize MAE-angle ($\circ$)  & \multicolumn{1}{c|}{6.24}   & \multicolumn{1}{c|}{16.1}    & 12.01           & 10.62                   \\ \hline
    \multirow{2}{*}{$\mathbf{t}_{pose}$}   & RMSE (m)                     & \multicolumn{1}{c|}{0.228}  & \multicolumn{1}{c|}{0.212}  & 0.346           & 0.221                  \\ \cline{2-6} 
                                   & \footnotesize MAE-absolute (m)          & \multicolumn{1}{c|}{0.08}  & \multicolumn{1}{c|}{0.073}  & 0.11           & 0.076                  \\ \hline\noalign{\hrule height 0.5pt}
    \end{tabular}   
         % \\ \vspace{2pt} * All rotation ($\mathbf{R}$) errors are measured in degrees, \\ and all translation ($\mathbf{t}$) errors are measured in meters (m).
         \vspace{-15pt}
    \end{table}

    % 본 실험에서는 키포인트를 기반으로 PnP solver를 이용해 3D 자세를 추정하였다. 
    % 기존 키포인트 검출 방법들은 정확도가 낮아 PnP solver 적용이 어려웠으며(표 X 참고), 이에 따라 우리 방법에 집중하여 실험을 진행하였다.
    % ♣ 여기 빠졌음! -> Finally, further
    We estimated the 3D pose using a PnP solver based on keypoints.
    Due to low accuracy (see Tab. \textcolor{blue}{\ref{tab:02}}), other methods were not considered for 3D pose reconstruction.
    Thus, we conducted a 3D pose estimation experiment focusing on the proposed DroneKey.
    As summarized in Tab. \textcolor{blue}{\ref{tab:06}}, the results indicate an average $\mathbf{R}_{pose}$ MAE-angle error of $10.625^\circ$.
    For $\mathbf{t}_{pose}$, the RMSE error was 0.221m, while the MAE-absolute error averaged 0.076m.
    Fig. \textcolor{blue}{\ref{fig:10}} provides a qualitative comparison of the 3D drone pose estimation results.

    In this experiment, rotation is visualized using only the drone camera's viewing axis for the clear comparisons.
    Additional quantitative results are provided in the supplementary materials.
    The results show that for all sequences, the estimated location and orientation closely match the ground truth.
    These findings confirm the effectiveness of our proposed method.

%■■■■■■■■■■■■■■■■■■■■■■■■■■■■■■■■■■■■■■■■■■■■■■■■■■■■■■■■■■■■■■■■■■■■■■■■■■■

\section{Conclusions}
\label{sec:conclusions}

This study proposes a drone-specific approach to address challenges in drone keypoint detection (Fig. \textcolor{blue}{\ref{fig:01}} (a)).
It enables 3D pose estimation using only geometric computations (e.g., PnP solver).
A transformer is employed to learn spatial relationships and accurately distinguish propellers with similar visual.
To enhance keypoint learning, we extract compact representations and apply gated sum in the keypoint head.
Additionally, a pose-adaptive Mahalanobis loss function improves training performance across diverse drone poses.
We released a synthetic drone keypoint dataset with real-world 360-degree backgrounds to address data shortages.
Using only 6 encoder layers, the proposed model achieved real-time performance at 44 FPS with high accuracy (OKS $AP^{kp}$ 99.68\%, RMSE 0.221m, MAE-absolute 0.076m, MAE-angle 10.62°).
The proposed DroneKey maintained high accuracy, demonstrating its suitability for real-time anti-drone systems.

Collecting real-world keypoint and 3D pose data is challenging, so we generated synthetic data with 360-degree images. 
Future study should focus on creating real-world datasets and handling drones with different propeller numbers. 
The proposed method estimates 3D pose using geometric calculations instead of a decoder. 
This emphasizes keypoint detection and enables robust pose estimation without extra training.
However, keypoint errors can distort 3D pose estimation.
To improve accuracy, we can consider adding a new 3D pose decoder in the future study. 
Since the proposed methods run in real time, adding a decoder is expected to improve stability while maintaining performance.

%■■■■■■■■■■■■■■■■■■■■■■■■■■■■■■■■■■■■■■■■■■■■■■■■■■■■■■■■■■■■■■■■■■■■■■■■■■■

% \section*{APPENDIX}
% The code and dataset are available at \url{https://github.com/kkanuseobin/DroneKey}.

%■■■■■■■■■■■■■■■■■■■■■■■■■■■■■■■■■■■■■■■■■■■■■■■■■■■■■■■■■■■■■■■■■■■■■■■■■■■

\section*{ACKNOWLEDGMENT}
{
% \small
\footnotesize
\linespread{0.5}\selectfont
This work was supported by the Institute of Information \& Communications Technology Planning \& Evaluation (IITP) grants funded by the Korea government (MSIT): the Artificial Intelligence Convergence Innovation Human Resources Development (IITP-2023-RS-2023-00256629) and the University ICT Research Center (ITRC) support program (IITP-2025-RS-2024-00437718). 
We appreciate the high-performance GPU computing support of HPC-AI Open Infrastructure via GIST SCENT.
}

%■■■■■■■■■■■■■■■■■■■■■■■■■■■■■■■■■■■■■■■■■■■■■■■■■■■■■■■■■■■■■■■■■■■■■■■■■■■
\linespread{0.8} % 기본값은 1, 0.9로 줄이면 간격이 좁아짐

\end{document}